\documentclass[10pt,journal,compsoc]{IEEEtran}
\usepackage{amssymb}
\usepackage{multirow}
\usepackage{booktabs}
\usepackage{algorithm}
\usepackage{algorithmic}
\usepackage{subfigure}
\usepackage{flushend}
\usepackage{ragged2e}
\usepackage{url}
\usepackage{threeparttable}
\usepackage{graphicx}

\usepackage{amsmath,amsfonts}
\usepackage{algorithmic}
\usepackage{algorithm}
\usepackage{array}
\usepackage[caption=false,font=normalsize,labelfont=sf,textfont=sf]{subfig}
\usepackage{textcomp}
\usepackage{stfloats}
\usepackage{verbatim}
\usepackage{graphicx}
\usepackage{cite}
\usepackage{bbm}
\newcolumntype{R}{>{\color{red}}c}

\usepackage{amsmath}
\title{\text{Diff-RNTraj: A Structure-aware Diffusion Model for} \text{Road Network-constrained Trajectory Generation}}

\author{Tonglong Wei, Youfang Lin, Shengnan Guo, Yan Lin, Yiheng Huang, \\Chenyang Xiang, Yuqing Bai, Huaiyu Wan
\thanks{~~Corresponding author: Shengnan~Guo}
\IEEEcompsocitemizethanks{\IEEEcompsocthanksitem T. Wei, Y. Lin, S. Guo, Y. Lin, Y. Huang, C. Xiang, Y. Bai and H. Wan are with School of Computer and Information Technology, Beijing Jiaotong University, Beijing 100044, China; And the Beijing Key Laboratory of Traffic Data Analysis and Mining, Beijing 100044, China. 
\protect\\
E-mail: \{weitonglong, yflin, guoshn, ylincs, huangyiheng, cyxiang, baiyuqing, hywan\} @bjtu.edu.cn }
\thanks{Manuscript received xx xx, xxxx; revised xx xx, xxxx.}
}

\newtheorem{definition}{Definition}

\usepackage{color}

\begin{document}

\IEEEtitleabstractindextext{%
\begin{abstract}
\justifying
\textcolor{black}{Trajectory data is essential for various applications. However, publicly available trajectory datasets remain limited in scale due to privacy concerns, which hinders the development of trajectory mining and applications. Although some trajectory generation methods have been proposed to expand dataset scale, they generate trajectories in the geographical coordinate system, posing two limitations for practical applications: 1) failing to ensure that the generated trajectories are road-constrained. 2) lacking road-related information.
In this paper, we propose a new problem, road network-constrained trajectory (RNTraj) generation, which can directly generate trajectories on the road network with road-related information. Specifically, RNTraj is a hybrid type of data, in which each point is represented by a discrete road segment and a continuous moving rate. To generate RNTraj, we design a diffusion model called Diff-RNTraj, which can effectively handle the hybrid RNTraj using a continuous diffusion framework by incorporating a pre-training strategy to embed hybrid RNTraj into continuous representations. During the sampling stage, a RNTraj decoder is designed to map the continuous representation generated by the diffusion model back to the hybrid RNTraj format. Furthermore, Diff-RNTraj introduces a novel loss function to enhance trajectory’s spatial validity. Extensive experiments conducted on two datasets demonstrate the effectiveness of Diff-RNTraj.}
\end{abstract}

}

\maketitle
\IEEEdisplaynontitleabstractindextext
\IEEEpeerreviewmaketitle

\IEEEraisesectionheading{\section{Introduction}\label{sec1}}
\IEEEPARstart{T}{he} widespread deployment of GPS devices in intelligent transportation systems (ITS) makes it convenient to record where a vehicle is. 
In this case, a sequence of ordered geographical coordinates of a vehicle is termed a trajectory.
Mining trajectory data can support the development of various downstream applications in ITS, such as user demand prediction~\cite{zhao2019predicting,wang2021gallat,zhang2023grey}, vehicle navigation~\cite{joshi2001new}, and personalized route recommendation~\cite{dai2015personalized,ge2019route,wang2021personalized}.
Despite the pivotal role of trajectory data, publicly available trajectory datasets are usually limited in scale. 
This is mainly due to privacy policies and public concern, including regulatory restrictions on trajectory data storage and sharing in many countries and areas, and the concern of users on personal location embedded in trajectories being exposed.
The available small-scale trajectory datasets hinder the study of trajectory data mining, since the state-of-the-art trajectory data mining solutions, especially deep-learning based solutions, usually require large-scale datasets to achieve optimal results.

To obtain large-scale trajectory data, the study on \textbf{\textit{trajectory generation}} has gained much attention~\cite{zhang2022factorized, wang2021large, chen2021trajvae, choi2021trajgail}. 
\textcolor{black}{It aims to analyze human potential spatial transfer patterns in cities and generate simulated GPS location sequences, which is vital for many trajectory-based applications.
For example, traffic simulation systems utilize the generated trajectory to simulate vehicle movement across road networks, thereby calculating high-fidelity traffic conditions that prompt the advancement of navigation systems~\cite{zhang2022factorized,choe2015trajectory}. Companies such as Uber leverage synthetic trajectories to expand the scale of trajectory datasets and develop larger models to link users and trajectories~\cite{gao2019deeptrip}.
Employing generated trajectory data for research also addresses privacy concerns, as it is devoid of personal information~\cite{jiang2023continuous}.}

\begin{figure}
	\centering
	\includegraphics[width=0.5\textwidth]{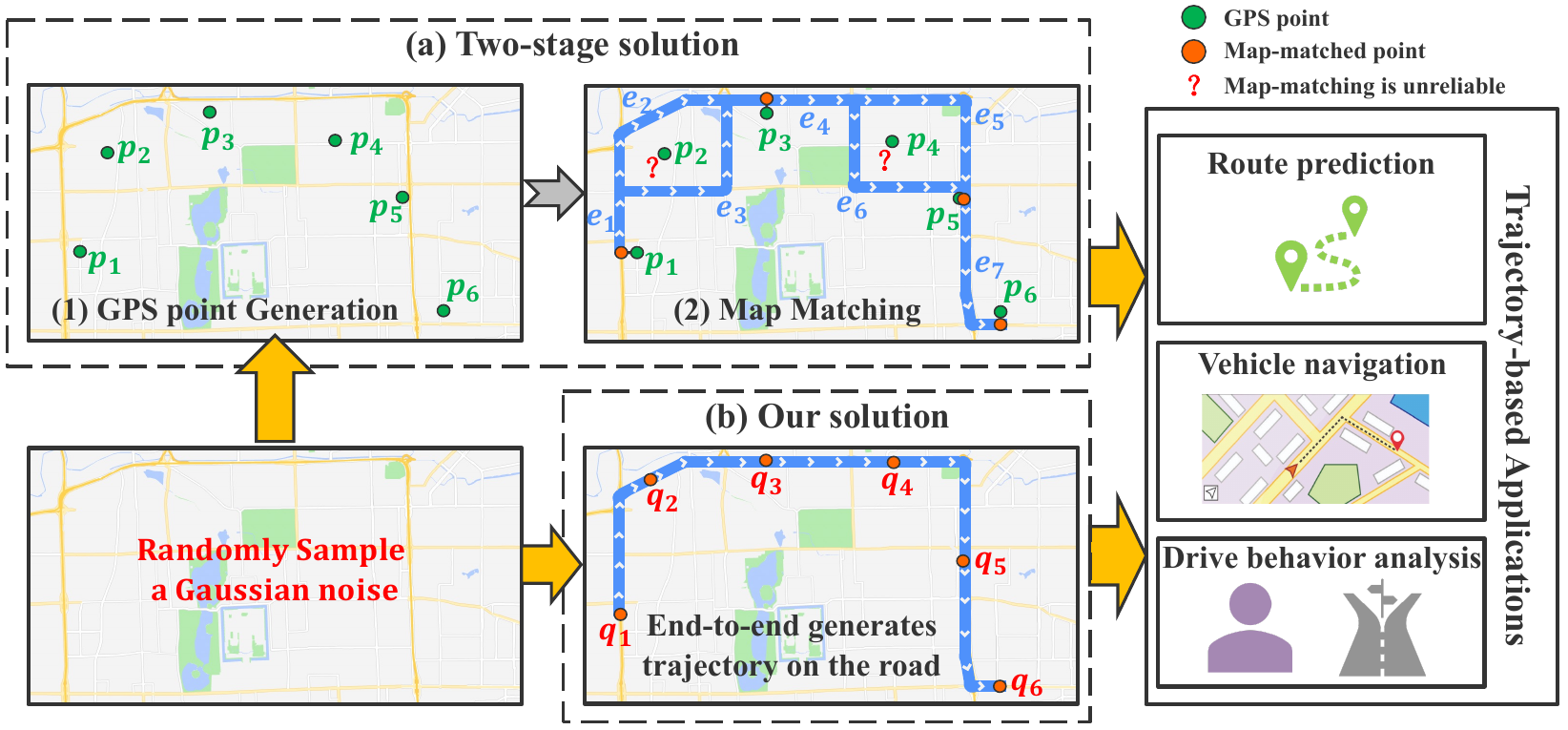}
	\caption{Comparison between the two-stage solution and our end-to-end solution. \textbf{Two-stage solution}: first generate GPS points, then perform map matching process. \textbf{Our solution}: directly generate trajectories on the road end-to-end.}
	\label{fig:end-to-end}
\end{figure}

Existing efforts for trajectory generation can be categorized into rule-based methods~\cite{isaacman2012human, zhao2019synthesizing} and deep learning methods~\cite{zhang2022factorized, long2023practical,ouyang2018non}.
\textbf{Rule-based methods} simulate human movement under the assumption that moving behavior adheres to predefined rules. 
However, they struggle to capture the complexity and uncertainty inherent in human travel patterns, limiting their ability to generate realistic trajectories.
By comparison, \textbf{deep learning methods} such as variational auto-encoder (VAE)-based methods~\cite{huang2019variational, kingma2019introduction} and generative adversarial network (GAN)-based methods\cite{smolyak2020coupled, goodfellow2014generative} are more flexible, since they are able to represent the complicated distribution of trajectories. 
However, all these methods represent trajectories by generating sequences of GPS points (\emph{i.e.}, latitude and longitude) in the geographical coordinate system, causing two limitations when the generated trajectories are utilized by downstream applications. Firstly, directly generating GPS points fails to ensure that the trajectories are on the road. As illustrated in Figure~\ref{fig:end-to-end}(a), the green GPS points $p_1 \sim p_6$ produced by current trajectory generation methods cannot accurately replicate the real moving patterns of vehicles on the road, as a valid and logical GPS point in the trajectory ought to stay on the road consistently.
Secondly, most downstream trajectory mining tasks, especially road-related applications, require not only precise GPS information (\emph{e.g.}, latitude and longitude) but also road-related information (\emph{e.g.}, the road segments on which the vehicle), yet such road-related information can not be retrieved directly from the generated GPS points. 

Therefore, to utilize the generated trajectory in downstream tasks, existing trajectory generation methods require an additional process, \emph{i.e.} applying map-matching~\cite{newson2009hidden} algorithm to the generated GPS sequences to ensure the final trajectories are valid and contain road-related information. 
However, such a two-stage strategy that first generates GPS points and then performs map matching is error-accumulating since the map-matching algorithm may return an unreliable result, especially when dealing with generated GPS points that significantly deviate from the road.
For example, in Figure~\ref{fig:end-to-end}(a), suppose that a vehicle's real travel path is $Path1: e_1 \to e_2 \to e_4\to e_5 \to e_7$. To reconstruct this trajectory, a two-stage strategy first generates six GPS points $\langle p_1, p_2, \cdots, p_6 \rangle$. Then, a map-matching algorithm is applied, which may map GPS points $p_2$, $p_4$ to road segments $e_3$ and $e_6$, respectively.
Finally, an undesired path $Path2: e_1 \to e_3 \to e_4\to  e_6 \to e_7$ is generated, deviating from the real one. 




Drawing on these observations, we propose a novel task called \textbf{\textit{road network-constrained trajectory generation}}, \textit{i.e.}, directly generating GPS trajectories on the road network in an end-to-end manner, thereby yielding trajectories directly applicable to subsequent applications. As shown in Figure~\ref{fig:end-to-end}(b), our newly proposed trajectory generation task not only ensures the validity of trajectory but also maintains detailed road-related information on vehicles' moving behavior. 
 Our proposed generation task calls for defining a new term \textbf{\textit{\underline{R}oad \underline{N}etwork-constrained \underline{Traj}ectory} (RNTraj)}.


\begin{figure}
	\centering
	\includegraphics[width=0.5\textwidth]{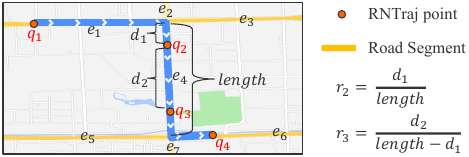}
	\caption{A RNTraj is a point sequence formed by $\langle q_1,q_2,q_3,q_4 \rangle$. Each RNTraj point $q$ can be determined by road segment $e$ and moving ratio $r$, \textit{e.g.}, the road segment of RNTraj point $q_2$ is $e_4$, and the moving ratio is $r_2$. For $q_3$, its road segment is $e_4$, and the moving ratio is $r_3$.}
	\label{fig:road_ratio}
\end{figure} 


A RNTraj $\tau_{RN}=\langle q_1, q_2, \cdots, q_T \rangle$ is a sequence of $T$ points in the road network coordinate system. Specifically, each RNTraj point $q = (e, r)$ is determined by a road segment $e$ on which the vehicle is traveling, and a moving ratio $r$ which indicates the proportion of a vehicle's newly traveled distance along the road segment to the initial remaining untraveled length on that segment. 
Figure~\ref{fig:road_ratio} shows an example of RNTraj. 
The road network-constrained trajectory generation task aims to generate trajectories in the form of RNTraj, ensuring that the generated trajectories are always on the roads and contain road-related information.

In this study, we propose to generate RNTraj based on diffusion model~\cite{sohl2015deep, ho2020denoising}, since diffusion model's capability of generating high-quality, realistic data have been proven in multiple domains, including image generation~\cite{rombach2022high, ho2022imagen,cao2024survey} and audio generation~\cite{liu2023audioldm, jeong2021diff}. 
\textcolor{black}{It demonstrates higher reliability and robustness compared to previous generative methods based on VAE and GAN~\cite{yang2022diffusion}.} 
\textcolor{black}{Furthermore, the diffusion model generates data from random Gaussian noise, which eliminates the risk of privacy leakage.}
Despite the promising performance of diffusion model, generating RNTraj with a diffusion model still faces the following unsolved challenges.

(1) \textbf{Hybrid data generation.} The state-of-the-art diffusion models can achieve superior performance in data generation, particularly for continuous data type such as image~\cite{hu2022global, ho2021classifier} and audio~\cite{chen2020wavegrad,kong2020diffwave}. However, in our case, RNTraj is ``\textit{hybrid}'', as it consists of a discrete road segment $e$ and a continuous moving ratio $r$. 
So existing diffusion models tailored toward the generation of a single type of data cannot be directly implemented for the generation of RNTraj.
Although employing two diffusion models to generate road segments and moving ratios separately is an option, it fails to consider the correlation between road segments and moving ratios in RNTraj. 
Furthermore, some studies~\cite{gao2022difformer,austin2021structured,he2022diffusionbert} indicate that the diffusion model for discrete data generation is still in the initial stage, and it faces challenges of producing high-quality results.

(2) \textbf{Spatial validity.} 
A generated trajectory is considered reasonable only when it adheres to fundamental physical principles and exhibits conventional driving behaviors. For example, in Figure~\ref{fig:road_ratio}, a reasonable trajectory can follow the route $ e_1\to e_4\to e_6$ rather than $e_1 \to e_6 \to e_3$, as road segment $e_6$ is disconnected to its predecessor $e_1$ and successor $e_3$.
However, diffusion models generate data through a multi-step denoising process, where each denoising step directly predicts the added noise, without taking into account the spatial validity of the final generated trajectories. Thus, how to incorporate spatial validity to guide trajectory generation in diffusion models requires consideration. 

To tackle the aforementioned challenges and achieve end-to-end RNTraj generation effectively, we propose a \textit{structure-aware \underline{diff}usion model for \underline{RNTraj} generation} (Diff-RNTraj). 
Specifically, Diff-RNTraj can handle hybrid data using continuous diffusion models by introducing a novel idea to convert a hybrid RNTraj into a continuous representation. This involves utilizing a pre-trained strategy to first learn the representation of discrete road segments, followed by concatenating the representation with the moving ratio together to obtain a vectorized RNTraj in the continuous space. 
Then, during the sampling stage, we propose a hybrid RNTraj decoder to map the generated continuous representations back to RNTraj format. Additionally, to ensure the spatial validity of the generated RNTraj, we incorporate a novel spatial validity loss at each step of the diffusion model’s denoising process to guide model training. 

In summary, the main contributions of this work are summarized as follows:
\begin{itemize}
    \item We propose a new problem, called road network-constrained trajectory generation, which directly generates trajectory on the road network, providing improved support for downstream trajectory-based applications compared to conventional trajectory generation.
    \item We design Diff-RNTraj to generate hybrid RNTraj by utilizing a continuous diffusion model. Diff-RNTraj integrates a pre-trained strategy to embed hybrid RNTraj into continuous representations, along with a decoder to convert the generated continuous representation back to hybrid RNTraj. Additionally, we introduce a novel spatial validity loss function to guarantee the spatial validity of the generated RNTraj.	
    \item Extensive experiments on two real-world trajectory datasets show that compared with other baselines, the trajectories produced by our Diff-RNTraj exhibit the highest level of rationality across a range of statistical metrics.
\end{itemize}

\section{Related Work}
\subsection{Trajectory Generation}
Generating large-scale, realistic trajectory data has garnered much attention in the research community. Early approaches~\cite{pfoser2003generating, theodoridis2000generating,pelekis2013hermoupolis} rely on predefined rules to simulate user behavior, disregarding the inherent randomness and complexity of human movement. Consequently, the generated trajectories fail to accurately capture real-world human movement patterns.

With the advance of generative models such as Variational Autoencoders (VAE)~\cite{kingma2019introduction,chen2021trajvae, zhang2022factorized, long2023practical} and Generative Adversarial Networks (GAN)~\cite{goodfellow2014generative, yu2017seqgan, ouyang2018non}, researchers have proposed a range of trajectory generation models that have shown promising results. TrajVAE~\cite{chen2021trajvae} follows the VAE framework, and utilizes Long Short-Term Memory networks (LSTM) to capture the temporal dependence and model trajectory data as a Gaussian distribution. It generates trajectory points step-by-step by sampling Gaussian noise and employing LSTM. EETG~\cite{zhang2022factorized} is a VAE-like model that effectively integrates both global and local trajectory semantic information to generate. 
SeqGAN~\cite{yu2017seqgan} employs the policy gradient algorithm for training. Its generator is implemented as an LSTM network and generates trajectories step by step. Furthermore, some works~\cite{feng2020learning,yuan2022activity,wang2023pategail} utilize generative adversarial imitation learning~\cite{ho2016generative} to generate trajectory.
In addition to directly generating GPS points, some researchers explore gridding trajectories~\cite{ouyang2018non, wang2021large}. They first employ a generative model to produce grid data, then convert it into trajectory data. DiffTraj~\cite{zhu2023difftraj} proposes a diffusion model method for trajectory generation, which first adds noise to the trajectory data and then removes the noise step-by-step to generate trajectory.
However, these methods focus on generating GPS points in the geographic coordinate system that require the map-matching process to project the generated GPS points onto the road before the generated trajectories are used. This two-stage solution will decrease the quality of generated trajectories since map-matching on the GPS points with errors is unreliable.

\textcolor{black}{There are some works that focus on generating the road-level trajectory~\cite{wang2022deep, jiang2023continuous, li2023trajectory}, which only generates the road segment sequences. Compared to the RNTraj generation task studied in this paper, road-level trajectory generation is coarse-grained. It cannot obtain precise latitude and longitude information for a vehicle, which will bring shortcomings when analyzing the trajectory behavior, e.g., the travel speed and acceleration patterns.}

\subsection{Diffusion Model}
The diffusion model is an iterative generative model that is proposed by~\cite{sohl2015deep} and further improved by~\cite{ho2020denoising, song2020score, song2020denoising}, which consists of two Markov processes: a forward diffusion process and a reverse denoising process. In the forward diffusion process, a clean sample is perturbed by gradually adding noise until it becomes a Gaussian noise. In the reverse denoising process, the added noise is removed step by step until a clean sample is obtained. 
Due to its high-quality generation capabilities, the diffusion model has gained widespread attention in the field of computer vision~\cite{rombach2022high, voleti2022mcvd} and natural language processing~\cite{li2022diffusion, liu2022diffsinger}.

According to the types of generated samples, diffusion models are divided into two types: continuous diffusion models, which are used for generating continuous data such as images~\cite{voleti2022mcvd} and speech data~\cite{kong2020diffwave}, and discrete diffusion models, which are used for generating discrete data like text~\cite{chen2022analog}. However, there exists a gap in utilizing diffusion models to directly generate hybrid data that combines both continuous and discrete data. For example, our RNTraj consists of discrete road segments and continuous moving rates, which is what our work focuses on.

\section{Preliminaries}
In this section, we first define the basic conceptions and then introduce the problem definition of road network-constrained trajectory generation.

\begin{definition}
[Trajectory] 
A trajectory $\tau$ is a sequence of GPS points, $\tau=\langle p_1, p_2, \cdots, p_T \rangle$, where $p_i = (lon_i, lat_i)$ represent the longtitude and latitude of $i$-th trajectory point, and $T = |\tau|$ is the length of trajectory.
\end{definition}

\begin{definition}
[Road Network] 
The road network is defined as a directed graph $\mathcal{G} = (\mathcal{V}, \mathcal{E})$, where $\mathcal{V}$ represents the set of the intersections between road segments, with each intersection having $lon$ and $lat$ attributes. $\mathcal{E}$ represents the set of road segments that link two intersections. Each road segment, denoted as $e \in \mathcal{E}$, has three attributes:
\begin{itemize}
	\item $start$: The start intersection of a road segment.
	\item $end$: The terminate intersection of a road segment.
	\item $length$: The length of a road segment in meters.
\end{itemize}
\end{definition}


\begin{definition}
[Road Network-constrained Trajectory]
We define a road network-constrained trajectory, \emph{i.e.}, RNTraj, of length $T$ as $\tau_{RN}$ = $\langle q_1, q_2, \cdots, q_T \rangle$, where each RNTraj point $q = (e, r)$ is located in the road coordinate system. Here, $e$ denotes the road segment where $q$ is located, and $r \in [0,1]$ is the moving ratio, representing the proportion of newly traveled distance covered along the road segment to the initial remaining untraveled length on that segment.
\end{definition}

Taking Figure~\ref{fig:road_ratio} as an example, we illustrate the meaning of the moving ratio $r$.
Here, we have a RNTraj form by $\langle q_1,\cdots, q_4 \rangle$ and points $q_2$ and $q_3$ both pass through road $e_4$, so their road segment are both $e_4$. For $q_2$, it is the first point to through road segment $e_4$, so its initial remaining untraveled length is equal to the total distance of road segment $e_4$. The distance it traveled on this road is from the start point of $e_4$ to the location of $q_2$, \textit{i.e.}, $d_1$. Therefore, the moving ratio of $q_2$ is calculated as $r_2 = \frac{d_1}{e_4.length}$. 
For $q_3$, it is not the first point to pass through this road. Consequently, its initial remaining untraveled length is the total length of the road segment minus the distance from the beginning of the road segment to its predecessor RNTraj point, \textit{i.e.}, $e_4.length - d_1$.  
The newly traveled distance on this road segment is $d_2$, and its moving rate is computed as $\frac{d_2}{ e_4.length - d_1}$. Therefore, we define the moving ratio $r_t$ for the $t$-th point $q_t$ in RNTraj as:	
\begin{equation}
\resizebox{0.91\linewidth}{!}{
\begin{math}
	r_t =   \begin{cases}
		\frac{dis(q_t.e.start, \, q_t)}{q_t.e.length}  & \text{if } q_t.e \neq q_{t-1}.e \text{ or } t=1 \\
		\frac{dis(q_{t-1}, \, q_t)}{q_t.e.length - dis(q_t.e.start, \, q_{t-1})} & \text{otherwise},
	\end{cases}
 \end{math}
 }
\end{equation}
where $dis(a,b)$ is the function to calculate the distance along the road network from $a$ to $b$. 

Given a RNTraj $\tau_{RN}=\langle q_1, \cdots, q_T \rangle$,  we can easily calculate the precise GPS coordinates of each RNTraj point according to Algorithm~\ref{alg:calGPS}.

\begin{algorithm}[htb]
	\caption{Calculate GPS Coordinate of Each RNTraj Point}
	\label{alg:calGPS}
	\textbf{Input}: A RNTraj $\tau_{RN}$ = $\langle q_1, \cdots, q_T \rangle$; \\   
	\textbf{Output}: The longitude and latitude of each RNTraj point;\\
        \textbf{Temporary variable}: $temp\_r$, used to calculate the final coordinates; \\
        \vspace{-\baselineskip}
	\begin{algorithmic}[1]  
        \FOR{$t = 1$ \textbf{to} $T$}
		\IF{$t = 1$ or $e_t \neq e_{t-1}$}
		\STATE $temp\_r \leftarrow r_t$; \\
		\ELSE
		\STATE $temp\_r \leftarrow temp\_r + (1 - temp\_r) * r_t$; \\
		\ENDIF
		\STATE $q_t.lon \leftarrow e_t.start.lon$ \\
		\STATE $\quad  \quad  \quad  \quad + (e_t.end.lon - e_t.start.lon) * temp\_r$; \\
		\STATE $q_t.lat \leftarrow e_t.start.lat$\\
		\STATE $\quad  \quad  \quad  \quad +(e_t.end.lat - e_t.start.lat) * temp\_r$;\\
        \ENDFOR
	\end{algorithmic}
\end{algorithm}

\begin{definition}
[Road Network Constrained Trajectory Generation]
Given a real-world trajectory dataset, the goal of this problem is to train a deep learning model, which can generate a RNTraj dataset that follows the distribution of original trajectory data.
\end{definition}

\begin{figure*}
	\centering
	\includegraphics[width=1\textwidth]{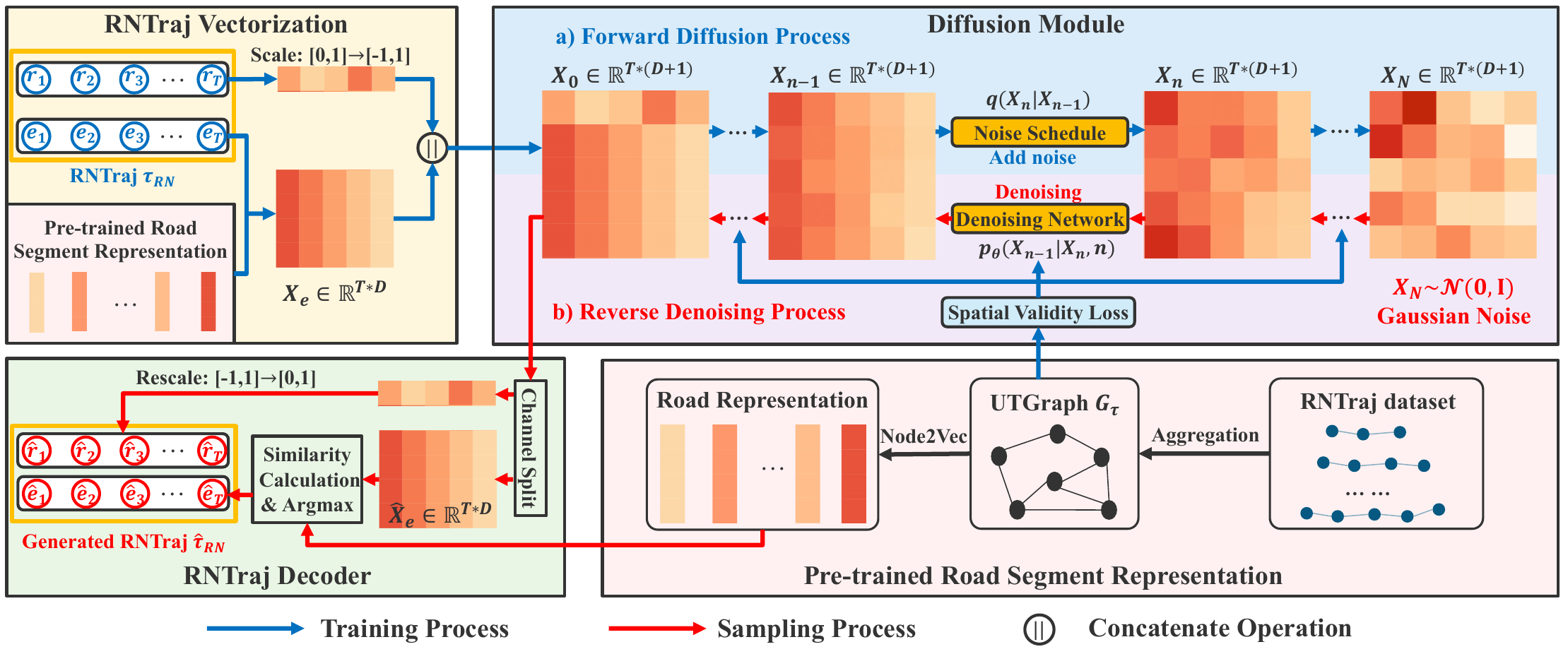}
	\caption{The architecture of Diff-RNTraj, which consists of a RNTraj vectorization module, a diffusion module, and a RNTraj decoder module.}
	\label{fig:model}
\end{figure*} 

\section{Methodology}
In this section, we introduce the Diff-RNTraj framework, which enables end-to-end generation of RNTraj using the diffusion model. The framework, depicted in Figure~\ref{fig:model}, is composed of three main modules: the RNTraj vectorization module, the diffusion module, and the RNTraj decoder module. 
The RNTraj vectorization module is responsible for converting a RNTraj from the hybrid format into a continuous representation space. This continuous representation serves as the input to the subsequent continuous diffusion model. 
The diffusion module learns the data distribution of RNTraj by applying a multi-step process of adding noise and denoising. 
Once the training process is completed, it allows sampling a vectorized RNTraj in continuous space from Gaussian noise. 
Finally, the generated continuous vectorized RNTraj is then mapped back to the hybrid format by the RNTraj decoder. 
In addition, to ensure that the generated trajectory is spatially valid, we incorporate a novel spatial validity loss function to train the Diff-RNTraj framework.
\subsection{RNTraj Vectorization}
\label{pre-train}
In order to leverage the prominent capabilities of the continuous diffusion model, it is essential to transform the hybrid RNTraj into a continuous representation. 
The core of this transformation is to convert the road segment from the discrete type into a continuous representation, meanwhile the continuous representations of the road segments should reflect the travel patterns of users on the road network. 
To achieve this, we propose a pre-training strategy for performing this conversion process.

Specifically, to effectively learn road segment representations, it is crucial to model the relationship between road segments. An intuitive approach to model these   relationships is by directly utilizing the topological structure of the road network. 
However, it is important to note that these topological relationships do not always align with the practical relationships observed in users' common travel patterns. For example, some physically connected roads are practically impassable due to traffic control measures. Besides, some road segments are frequently traveled successively by users (e.g., major roads), while others are not. Therefore, the weights between the road segments are different and cannot be accurately represented by the road network topology.

To this end, we introduce UTGraph, a road segment connectivity graph constructed from user trajectories. UTGraph directly establishes connections between road segments based on real user trajectories. It is defined as $\mathcal{G}_\tau = (R, \mathcal{E}_\tau)$, where $R$ is the set of nodes representing road segments, and $\mathcal{E}_\tau$ is the set of edges representing transition frequencies between road segments, which can be extracted from road network constrained trajectory as follows.
\begin{itemize}
	\item For nodes $r_i, r_j \in R$, there exists an edge $e_{ij} \in \mathcal{E}_\tau$  if there is a RNTraj consecutively passing road segment $i$ and road segment $j$.
	\item The weight of edge $e_{i,j}$ is the total number of times that road segments $i$ and $j$ are consecutively traversed in all trajectory data.
\end{itemize}

After constructing UTGraph, we employ the Node2vec algorithm~\cite{grover2016node2vec} to calculate the road segment representation $\textbf{E} \in \mathbb{R}^{|R| \times D}$, where $D$ is the dimension of the road segment representation, $|R|$ is the number of road segments.

By combining the road segment representation and the moving ratio along the channel dimension, we can represent a RNTraj $\tau_{RN}$ with length $T$ as a tensor $X\in \mathbb{R}^{T \times (D+1)}$, which is referred to as vectorized RNTraj. The first $D$ dimensions of the vectorized RNTraj correspond to road segment representations, denoted as $X_e \in \mathbb{R}^{T \times D}$, and the last dimension represents the moving rate $r$. 
To scale the moving rate within the range $[-1, 1]$, we use the transformation $X{[:, D+1]} = 2 * r - 1$. 

With this approach, we can represent a RNTraj in continuous space. Next, we employ the continuous diffusion model to generate the vectorized RNTraj.

\subsection{Diffusion Model for Vectorized RNTraj Generation}



To learn a distribution $p_{\theta}(X)$ that approximates the continuous distribution of vectorized RNTraj $q(X)$ and to allow easy sampling, we utilize the diffusion model framework. This framework encompasses two key steps: the forward diffusion process and the reverse denoising process.

\subsubsection{The Forward Diffusion Process} 
The forward diffusion process gradually introduces noise to the sample until the sample follows the standard Gaussian distribution. Suppose that we have an original vectorized RNTraj, $X_0$, the forward diffusion model aims to obtain a noisy vectorized RNTraj, denoted as $X_N$, through $N$ steps forward process. This process can be defined as the following Markov chains:
\begin{equation}
	q(X_{N} \mid X_0) = \prod_{n=1}^{N} q(X_n \mid X_{n-1}),
\end{equation}
where the forward diffusion process $q(X_n \mid X_{n-1})$ at each step follows a Gaussian distribution:
\begin{equation}
	\label{onestep_xn}
	q(X_n \mid X_{n-1}) = \mathcal{N}(X_n; \sqrt{(1 - \beta _n)}  X_{n-1}, \beta _n  \text{I}),
\end{equation}
where $\text{I}$ represents the identity matrix. $\beta_n \in (0,1)$ is fixed while $\beta_1, \beta_2, \cdots, \beta_N$ gradually increase to control the level of noise added in each forward step.

Since $\beta_n$ is fixed at each step, we can simplify Equation~\ref{onestep_xn} by utilizing the properties of the Gaussian process to derive the distribution of $X_n$ given $X_0$ for each step as:
\begin{equation}
	q(X_n\mid X_0) = \mathcal{N}(X_n; \sqrt{\bar{\alpha }_n }X_0, (1-\bar{\alpha }_n) \text{I}),
\end{equation}
where $\bar{\alpha } _n = \prod_{i=1}^{n}\alpha _i $, and $\alpha _i = 1 - \beta _i$. By utilizing the reparameterization technique~\cite{kingma2013auto}, we can directly compute $X_n$ by: 
\begin{equation}
	\label{onestep}
	X_n = \sqrt{\bar{\alpha }_n }X_0 + \sqrt{(1 - \bar{\alpha }_n) }\epsilon , 
\end{equation}
where $\epsilon \sim \mathcal{N}(0,\text{I})$.

Finally, the noisy vectorized RNTraj in the last step of the forward diffusion process follows the standard Gaussian distribution.
\begin{equation}
	\label{gausian}
	X_N \sim \mathcal{N}(X_N; 0, \text{I} )
\end{equation}

\subsubsection{The Reverse Denoising Process}
The objective of the reverse denoising process is to transform $X_N$ into a noise-free vectorized RNTraj by gradually removing noise step-by-step. Each step of the reverse denoising process is defined as:	
\begin{equation}
	\label{denoise}
    p(X_{n-1} \mid X_n) = \mathcal{N}(X_{n-1}; \mu _{n-1},  \textstyle \sum_{n-1} ),
\end{equation}
where $\mu _{n-1}$ and $\textstyle \sum_{n-1}$ denote the mean and variance at the $(n-1)$-th step, respectively. If the values of mean and variance at each step are known, $X_{n-1}$ can be sampled directly. However, these values are unknown in practice. Therefore, we employ a neural network with parameters $\theta$ for estimating the mean and variance. The Equation~\ref{denoise} can be reformulated as follows.
\begin{equation}
	p_{\theta }(X_{n-1} \mid X_n) = \mathcal{N}(X_{n-1}; \mu_{\theta}(X_n,n),  \textstyle \sum_{\theta}(X_n,n)),
\end{equation}
where $\mu_{\theta}(X_n,n)$ and $\textstyle \sum_{\theta}(X_n,n)$ are the learned mean and variance.
Following DDPM~\cite{ho2020denoising}, we predict the added noise at each step of the forward diffusion process rather than directly calculating the mean. Therefore, we re-parameterize the mean $\mu_{\theta}(\cdot)$ and $\textstyle \sum_{\theta}(\cdot)$ as follows.
\begin{equation}
\begin{split}
    \mu _{\theta }(X_n,n) &= \frac{1}{\sqrt{\alpha _n} }(X_n - \frac{\beta _n}{\sqrt{1-\bar{\alpha }_n } }\epsilon _{\theta }(X_n,n)) \\
    {\textstyle \sum_{\theta }}(X_n,n) &= \frac{1-\bar{\alpha }_{n-1}}{1-\bar{\alpha }_{n}}\beta _n,
\end{split}
\label{eq:re-parameterize-mean-var}
\end{equation}
where $\epsilon _{\theta }(\cdot)$ is the denoising network that is used to estimate the amount of noise to be removed at the current step. And $\epsilon _{\theta }(X_n,n)$ is the estimated noise.

The complete reverse denoising process to infer a vectorized RNTraj can be formulated as follows:
\begin{equation}
	p(X_0 | X_N) = p(X_N) \prod_{n=N}^{1}p(X_{n-1} | X_n),
\end{equation}

Since $X_N$ follows a standard Gaussian distribution, as defined in Equation~\ref{gausian}, we can directly get a clean vectorized RNTraj $\hat{X}_0$ in the continuous space from a random Gaussian noise by the reverse denoising process of $N$ steps.

\begin{figure}
	\centering
	\includegraphics[width=1.0\linewidth]{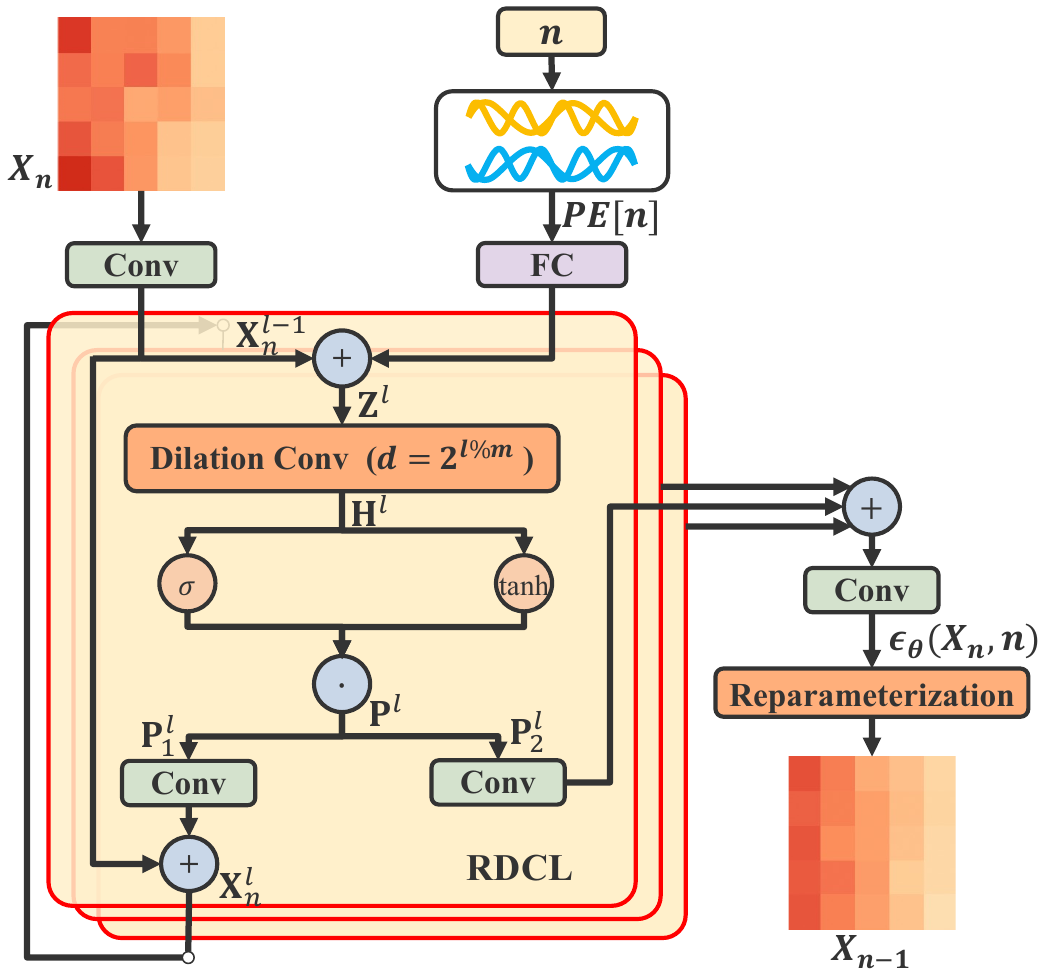}
	\caption{The architecture of the denoising network, which is stacked by $L$ residual dilation convolution layers (RDCLs).}
	\label{fig:denoise}
\end{figure} 

\subsubsection{Denoising Network}
The denoising network $\epsilon_\theta(\cdot)$ in Equation~\ref{eq:re-parameterize-mean-var} takes the noisy vectorized RNTraj $X_n$ and the step $n$ as input to predict the added noise, which can be used to calculate $X_{n-1}$.
Generally, the denoising network is designed as a Wavenet-based architecture for sequence-related tasks~\cite{liu2022diffsinger, kong2020diffwave, kim2020glow}, using convolution operation to capture the temporal dependence of sequences. Following DiffWave~\cite{kong2020diffwave}, we utilize the residual dilation convolutional layer as the basic component of the denoising network to capture temporal correlations of trajectory data and expand the receptive field.

Specifically, the framework of the denoising network is shown in Figure~\ref{fig:denoise}. At the $n$-th step of the denoising process, we first implement the positional encoding proposed in DiffWare~\cite{kong2020diffwave} to project the step $n$ into a $F$-demensional vector $PE[n] \in \mathbb{R}^F$:
\textcolor{black}{\begin{equation}
	\begin{split}
		PE[n] = [&sin(10^{\frac{0 * 4}{F / 2}}n),sin(10^{\frac{1 * 4}{F / 2}}n),\cdots ,sin(10^{\frac{(F/2-1) * 4}{F / 2}}n),\\
		&cos(10^{\frac{0 * 4}{F / 2}}n),cos(10^{\frac{1 * 4}{F / 2}}n),\cdots ,cos(10^{\frac{(F/2-1) * 4}{F / 2}}n)],
	\end{split}
\end{equation}
and we implement two fully connected layers on the vector.}
To capture the temporal correlation within trajectory data, we stack $L$ Residual Dilation Convolution Layers~(RDCL). These layers are grouped into $\frac{L}{m}$ blocks, each consisting of $m$ layers. 

The input to the $l$-th RDCL consists of two parts, including the output $\textbf{X}^{l-1}_n \in \mathbb{R}^{T\times C}$ of the $(l-1)$-th layer, and the positional encoding of $n$. For the first RDCL, $\textbf{X}_n^0$ is transformed from sample $X_n$. Formally, 
\begin{equation}
\begin{split}
    \textbf{Z}^l &= \textbf{X}_n^{l-1} + \mathrm{FC}(PE[n]) \in \mathbb{R}^{T \times C} \\
    \mathrm{FC}(\cdot) &: \mathbb{R}^{F} \to \mathbb{R}^{C},
\end{split}
\end{equation}
where $C$ represents the number of channels in RDCL. 
Then, we input it into a dilation convolution to capture the temporal dependencies at various scales by setting the dilation rate $d=2^{l\%m}$. The dilation rate doubles at each layer within a block, \textit{i.e.}, $[1, 2, \dots, 2^{m-1}]$. Subsequently, the output of the dilation convolution $\textbf{H}^l \in \mathbb{R}^{T \times 2C}$ is fed into a gate unit to filter out useful information:
\begin{equation}
	\textbf{P}^l = \sigma (\textbf{H}^l) \odot  \mathrm{tanh}(\textbf{H}^l) \in \mathbb{R}^{T \times 2C},
\end{equation}
where $\sigma(\cdot)$ denotes the sigmoid function and $\odot $ is element-wise multiplication. $\textbf{P}^l$ is further split into $\textbf{P}_1^l$ and $\textbf{P}_2^l$ along the channel dimension:
\begin{equation}
    \begin{split}
        \textbf{P}_1^l  &= \textbf{P}^l{[:, 0:C]}\in \mathbb{R}^{T\times C} \\
        \textbf{P}_2^l  &= \textbf{P}^l{[:, C:2C]}\in \mathbb{R}^{T\times C}
    \end{split}
\end{equation}

We take $\textbf{P}_1^l$ and the input of current layer $\textbf{X}^{l-1}_n$ through residual connections as the output of $l$-th layer, \textit{i.e.}, $\textbf{X}_n^l = \textbf{X}_n^{l-1} + \mathrm{Conv}(\textbf{P}_1^l)$.
Finally, we sum $\textbf{P}_2^l$ of each layer to output the estimated noise, \textit{i.e}.,
\begin{equation}
    \epsilon_{\theta}(X_n, n) = \mathrm{Conv}( {\textstyle \sum_{i=1}^{L}\mathrm{Conv}(\textbf{P}_2^l ) } ) \in \mathbb{R}^{T \times (D+1)}
\end{equation}

Once we get the estimated noise, we can re-parameterize $X_{n-1}$ by:
\begin{equation}
	\label{cal_xn_1}
	\begin{split}
		X_{n-1} &= \frac{1}{\sqrt{\alpha _n} }(X_n - \frac{\beta _n}{\sqrt{1-\bar{\alpha }_n } }\epsilon_{\theta }(X_n,n))  \\
		&+ \sqrt{\frac{1-\bar{\alpha }_{n-1}}{1-\bar{\alpha }_{n}} \beta _n}
	\end{split}
\end{equation}

\begin{algorithm}[t]
        \caption{Sampling of Diff-RNTraj}
        \label{infer_RNTraj}
	\textbf{Input}: Pre-trained road segment representations \textbf{E}, trained denoising network $\epsilon_{\theta}(\cdot)$, the length of RNTraj $T$, the total of diffusion steps $N$; \\   
	\textbf{Output}: A generated RNTraj; 
        \begin{algorithmic}[1]  
            \STATE Sample $X_N \sim \mathcal{N}(\textbf{0}, \textbf{I})$ where $X_N \in \mathbb{R}^{T \times (D+1)}$; \\
            \FOR{$n = (N-1)$ \textbf{to} $0$}
                \STATE Calculate $X_n \in \mathbb{R}^{T \times (D+1)}$ using Equation~\ref{cal_xn_1}; \\
            \ENDFOR
            \STATE Split generated $\hat{X}_0$ along the channel dimension, obtain RNTraj road segment representation $\hat{X}_e \in \mathbb{R}^{T \times D}$ and moving ratio $\hat{r}$ according to Equation~\ref{split_x0}; \\
            \STATE Calculate cosine similarity between $\textbf{E}$ and $\hat{X}_e$ use Equation~\ref{cos_sim}, and use argmax function to obtain generate road segment sequence $\hat{e}_1, \hat{e}_2, \cdots, \hat{e}_T$; \\
            \STATE Obtain the GPS coordinate of RNTraj according to Algothrim~\ref{alg:calGPS}.
        \end{algorithmic}
\end{algorithm}

\subsection{RNTraj Decoder}
\label{hybrid_decoder}
In order to convert the sampled vectorized RNTraj $\hat{X}_0 \in \mathbb{R}^{T \times (D+1)}$ into discrete road segments $\hat{e}$ and continuous moving ratio $\hat{r}$, we split $\hat{X}_0$ along the channel dimensions to obtain the road segment representation $\hat{X}_e$ and the moving ratio $\hat{r}$:
\begin{equation}
    \label{split_x0}
    \begin{split}
        \hat{X}_e &= \hat{X}_0{[:, 0:D]} \in \mathbb{R}^{T \times D} \\
        \hat{r} &= \frac{\hat{X}_0[: , D+1] + 1}{2} \in \mathbb{R}^{T \times 1}
    \end{split}
\end{equation}

To determine the road segment $\hat{e}$, we first calculate the cosine similarity between the pre-trained road segment representations $\textbf{E}$ and $\hat{X}_e$,
\begin{equation}
\label{cos_sim}        
    S = \frac{\hat{X}_e \cdot \textbf{E}^\top  }{\left \| \hat{X}_e \right \|_2 \cdot  \left \| \textbf{E} \right \|_2} \in \mathbb{R}^{T \times |R|},
\end{equation}
Then, we use argmax function to obtain road segment $\hat{e}_t$ of $t$-th RNTraj point, \textit{i.e.}, $\hat{e}_t = \mathop{\rm argmax}\limits_{i} S[t,i]$.

The generation process of RNTraj can be found in Algorithm~\ref{infer_RNTraj}.
Next, we will detail how to train the parameters $\theta$ of $\epsilon_{\theta}(\cdot)$ in the next section.

\subsection{Model Training}
To train the parameters $\theta$, DDPM employs the following loss function to minimize the error between the added noise $\epsilon$ and the estimated noise.
\begin{equation}
	\label{ddpm}
	 \mathcal{L}_1(\theta )=  \mathbb{E} _{{n, X_0, \epsilon}} ||\epsilon -\epsilon _\theta (\sqrt{\bar{\alpha}_n}X_0 + \sqrt{1-\bar{\alpha}_n }\epsilon, n)||^2
\end{equation}

However, this loss function only considers the added noise for guiding model training, and fails to take the structure information of RNTraj into consideration, which led to the generated RNTraj being invalid.
Inspire by~\cite{li2022diffusion}, at each step of the reverse denoising process, we further use the estimated vectorized RNTraj $\hat{X}_0$ based on the estimated noise
to fit the real vectorized RNTraj $X_0$, since $X_0$ contains the structure of the sample. Formally,

\begin{equation}
	\label{calx0}
	\mathcal{L}_2 =  ||X_0 - \hat{X}_0 ||^2,
\end{equation}
where $	\hat{X}_0 = (X_n - \sqrt{1 - \bar{\alpha }_n }\epsilon _\theta(X_n,n)   ) / \sqrt{\bar{\alpha }_n } $, as described in Equation~\ref{onestep}.


Furthermore, in order to consider sequence dependency and enhance the spatial validity of the generated trajectories, we propose a novel spatial validity loss function that explicitly models the connectivity of generated road segments. Based on the estimated $\hat{X}_0$, we reconstruct RNTraj following Section~\ref{hybrid_decoder} and obtain a generated road segment sequence $\langle \hat{e}_1, \hat{e}_2, \cdots, \hat{e}_T\rangle$.
To ensure that the adjacent road segments are connected in the UTGraph $\mathcal{G}_\tau$, we penalize cases where they are not connected, as these two road segments would be unreachable.
Hence, our loss function is defined as:
\begin{equation}
	\mathcal{L}_3  =  \sum_{i=1}^{T-1} (1 - \mathbbm{1}\{\mathcal{G}_\tau(\hat{e}_i,\hat{e}_{i+1} )\}),  
\end{equation}
where $\mathbbm{1}\{\cdot\}$ is an indicator function that $\mathbbm{1}\{\mathcal{G}_\tau(\hat{e}_i,\hat{e}_{i+1} ) \}= 1$ if the weight of $\hat{e}_i$ and $\hat{e}_{i+1}$ in $\mathcal{G}_\tau$ not equals 0; Otherwise, $\mathbbm{1}\{\mathcal{G}_\tau(\hat{e}_i,\hat{e}_{i+1} ) \}= 0$. 

Overall, our total loss function is defined as:
\begin{equation}
		\mathcal{L} = 	\mathcal{L}_1 + 	\mathcal{L}_2 +	\mathcal{L}_3
\end{equation}
The detailed training algorithm is presented in Algorithm~\ref{training_algorithm}.

\begin{algorithm}[tb]
        \caption{Training of Diff-RNTraj}
	\label{training_algorithm}
	\textbf{Input}: Pre-trained road segment representation \textbf{E}, the distribution of train dataset, the total forward diffusion step $N$, the variable schedule $\{\beta_1,\cdots,\beta_N\}$, UTGraph $G_\tau$ ;\\   
    \textbf{Output}: Trained denoising network $\epsilon_\theta(\cdot)$;\\
            \vspace{-\baselineskip}
 \begin{algorithmic}[1]  
            \WHILE{model not converged}
             \STATE Sample a RNTraj, calculate the corresponding vectorized RNTraj $X_0$ in continuous space; \\
		 \STATE Sample $n \sim \mathrm{Uniform}(\{1, \cdots ,N\})$, $\epsilon \sim \mathcal{N}(0,\textbf{I})$; \\
		 \STATE Calculate $X_n$ according to Equation~\ref{onestep}; \\
		 \STATE Calculate the estimated $\epsilon_\theta(X_n, n)$, $\hat{X}_0$, and reconstruct RNTraj; \\
		 \STATE Calculate the loss $\mathcal{L} = \mathcal{L}_1 + \mathcal{L}_2 + \mathcal{L}_3$; \\
		 \STATE Take the gradient step;
            \ENDWHILE
        \end{algorithmic}
\end{algorithm}

\section{Experiments}
In this section, we conduct extensive experiments to evaluate the quality of the generated trajectory by the proposed model Diff-RNTraj on two real-world trajectory datasets and compare it with other baselines.

\begin{table}[h]	
	\small
	\centering
	\caption{Dataset description.}
 \resizebox{1.0\linewidth}{!}{
	\begin{tabular}{c|c|c}
		\toprule
		Types & Porto & Chengdu \\
		\midrule
		Latitude & (41.1405, 41.1865) & (30.6550, 30.7291) \\
		Longtitude & (-8.6887, -8.5557) & (104.0393, 104.1271) \\
		\# Trajectory & 537,681 & 527,353 \\
		\# Road segment & 13695 & 6256 \\
		\# Time interval & 15s & 30s\\
		Mean trajectory length & 41.50 & 29.81\\
		\bottomrule
	\end{tabular}
 }
	\label{tab:dataset}
\end{table}

\subsection{Dataset}
In experiments, we use two publicly available trajectory datasets from Porto, Portugal, and Chengdu, China. The Porto dataset\footnote{\url{https://www.kaggle.com/competitions/pkdd-15-predict-taxi-service-trajectory-i/data}} contains trajectory data of 442 taxis from January 2013 to June 2014. The Chengdu dataset\footnote{\url{https://outreach.didichuxing.com/}} records taxi trajectory data during October 2016. We download the road network from OpenStreetMap\footnote{\url{http://www.openstreetmap.org/}} using osmnx tools. 
We remove the stationary points and filter out trajectories that traveled less than 5 minutes or more than 30 minutes. Then, we apply the map matching algorithm~\cite{newson2009hidden} to project GPS points onto the road network and obtain the ground truth of RNTraj.
Table~\ref{tab:dataset} provides statistical information of datasets after pre-processing.
\textcolor{black}{Since we used datasets with fixed time intervals between adjacent GPS points, and our generated trajectories also obey the corresponding time patterns, which is helpful for further analysis of trajectory behavior, e.g., varying speeds and acceleration patterns.}

\subsection{Evaluated Metrics}
As our goal is to generate a realistic synthetic trajectory dataset that can benefit various trajectory mining tasks. Hence, it is essential to evaluate the ``similarity" between the generated trajectory dataset and the real dataset from a macro perspective. Following~\cite{wang2023pategail, long2023practical}, we use the Jensen-Shannon Divergence (JSD) to quantify the disparity between them. For two distributions $p$ and $q$, the JSD between them can be calculated by:
\begin{equation}
	\text{JSD}(p,q) = \frac{1}{2}\text{KL}(p||\frac{p+q}{2}) + \frac{1}{2}\text{KL}(q||\frac{p+q}{2}),
\end{equation}
where $\text{KL}(\cdot||\cdot)$ is the Kullback-Leibler divergence~\cite{cover1999elements}. A smaller JSD value indicates a higher similarity between distributions $p$ and $q$. We assess the quality of the generated trajectories by calculating the JSD with the following metrics:

\begin{itemize}
	\item \textbf{Trajectory Total Distance (JSD-TD)}  is used to calculate the distribution of the total distance of a trajectory. 
	\item \textbf{Trajectory Segment Distance (JSD-SD)} is used to evaluate the travel distance in trajectory segments, \textit{i.e.}, the length between two adjacent trajectory points.
	\item \textbf{Grid Point Density (JSD-GPD)} measures the distribution of all trajectory points. Specifically, we divide the map into multiple grids with grid size 50m $\times$ 50m, and we count the probability that the trajectory point is located in each grid.
	\item \textbf{Road Segment (JSD-RS)} quantifies the road segment distribution of our generated trajectory.
\end{itemize}

Besides evaluating the overall trajectory distribution, we design a novel metric, called \textbf{Road Segment Connectivity (RSC)}, to assess the spatial validity of the generated trajectories. RSC measures the reachable proportion of adjacent points within generated trajectories. We determine the road segment connectivity by UTGraph $\mathcal{G}_\tau$ since it is reflected by the real vehicle trajectory. For a generated RNTraj $\hat{\tau}_{RN} = \langle \hat{q}_1, \cdots, \hat{q}_T \rangle$, its RSC is calculated as follows:
\begin{equation}
	\textbf{RSC} = \frac{ {\textstyle \sum_{t=1}^{T-1}} \mathbbm{1}\{\mathcal{G}_\tau (\hat{q}_t.e, \, \hat{q}_{t+1}.e) \}}{ { T - 1} } \times 100\%
\end{equation}
A higher RSC value signifies a higher reasonableness in generated trajectories. 

\subsection{Baselines}
We compare our proposed model with the following baselines:
\begin{itemize}
	\item \textbf{Random Walk on the Road Network (RWRN).} RWRN randomly selects a point as the origin, then performs the random walk on the road network according to a random speed and direction.
        \item \textcolor{black}{\textbf{Markov~\cite{gambs2012next}}. Markov treats the road segments as states and constructs a transition matrix to capture the first-order transition probabilities between these road segments.}        
	\item \textbf{TrajVAE~\cite{chen2021trajvae} + Hidden Markov Model (HMM)~\cite{newson2009hidden}.} TrajVAE learns the trajectory representation using LSTM and VAE, and reconstructs the GPS points through a decoder. Then, a map-matching algothrim based on the HMM is implemented to project GPS points into the road network.
        \item  \textbf{TrajGAN~\cite{yu2017seqgan} + HMM.} TrajGAN employs an LSTM network in its generator to generate GPS points step by step, which is trained using the policy gradient algorithm. Subsequently, these GPS points are projected onto the road network based on HMM. 
	\item \textbf{EETG~\cite{zhang2022factorized} + HMM~\cite{newson2009hidden}.} EETG is a VAE-like trajectory generate method, which combines VAE and VRNN~\cite{chung2015recurrent} to encode trajectory, and generate trajectory based on LSTM. Then, HMM is introduced to obtain the road network-constrained trajectory.
\end{itemize}
In addition, we design the following variants to demonstrate the effectiveness of the components in the proposed model.
\begin{itemize}
        \item \textbf{Diff-Bit.} We utilize the binary encoding to represent the discrete road segment~\cite{chen2022analog} and concatenate the moving ratio to vectorize hybrid RNTraj. During the sampling stages, we decode the discrete road segment by threshold operations. 
	\item \textbf{Diff-RNTraj / Pretrain.} Instead of using pre-trained methods, we randomly initialize the representation for each road segment and optimize it together with Diff-RNTraj.
	\item \textbf{Diff-Transformer.} We use the Transformer in the denoising network to capture the temporal correlation of the noisy vectorized RNTraj. Compared with our residual dilation convolution layers, Diff-Transformer considers the global correlation between all trajectory points.
        \item \textbf{Diff-LSTM.} We replace the residual dilation convolution layers in the denoising network with two LSTM layers, which only capture the temporal dependence among previous trajectory points.
	\item \textbf{Diff-RNTraj / SL.} We remove the spatial vality loss function, and only use $\mathcal{L}_1$ loss to train Diff-RNTraj.
\end{itemize}
\begin{table*}[t]
	\centering
	\caption{Performance comparison of our model and baselines on two real-world trajectory datasets. \\ \text{The best results are highlighted in \textbf{bold}, while the \underline{underline} indicates the second-best results. $\dag$ represents our model.}}
	\resizebox{1.0\linewidth}{!}{
		\begin{tabular}{c|ccccc|ccccc}
			\toprule
			\multirow{2}{*}{Methods}&\multicolumn{5}{c|}{Porto} &\multicolumn{5}{c}{Chengdu}\\
			\cmidrule{2-11}
			 & JSD-TD & JSD-SD & JSD-GPD  & JSD-RS & RSC  & JSD-TD & JSD-SD & JSD-GPD & JSD-RS & RSC \\
			\midrule
			RWRN & 0.6058 & 0.0149  & 0.1724  & 0.9206 & \textbf{100\%}  & 0.4267  & 0.0023  & 0.5819 & 0.9108 & \textbf{100\%} \\
                \textcolor{black}{Markov} & 0.5650 & 0.0122 & 0.1066 & \underline{0.6143} & \textbf{100\%}  & 0.3269 & 0.0019 & 0.1539 & \underline{0.4433} & \textbf{100\%} \\                
			TrajVAE + HMM & 0.3399 & 0.0093 & \underline{0.0791} & 0.7251 & 13.30\% & 0.3925 &  0.0026 & 0.1666 & 0.5662 & 6.95\% \\
                TrajGAN + HMM & 0.3876 & 0.0123 & 0.0885 & 0.7520 & 30.07\% & 0.4602 & 0.0029 & 0.1902 & 0.8868 & 6.43\%\\
			EETG + HMM & \underline{0.1956} & \underline{0.0014} & 0.0963 & 0.8652 & 23.83\% & \underline{0.1773} & \underline{0.0015} & \underline{0.1489} & 0.5170 & 18.09\%\\
                \midrule
			Diff-Bit & 0.1277 & 0.0025 & 0.0377 & 0.0868 & 27.63\% & 0.1397 & 0.0021 & 0.1028 & 0.0591 & 37.34\% \\
			Diff-RNTraj / PreTrain & 0.0975 & 0.0015 & 0.0449 & 0.2083 & 25.40\% & 0.1279 & 0.0018 & 0.0891 & 0.2544 & 48.87\% \\
                Diff-LSTM & 0.0512 & 0.0013 & 0.0372 & 0.1396 & 56.68\% & 0.1381 & 0.0018 & 0.0934 & 0.0819 & 71.41\% \\
			Diff-Transformer & 0.0579 & 0.0016 & 0.0356 & 0.0804 & 85.86\% & 0.1247 & 0.0012 & 0.0946 & 0.0544 & 86.75\% \\
			Diff-RNTraj / SL& 0.0483 & 0.0008 & 0.0300 & 0.0614 & 79.68\% & 0.1658 & 0.0015 & 0.0952 & 0.0467 & 87.79\%\\
                \midrule
			\textbf{Diff-RNTraj$\dag$} & \textbf{0.0221} &  \textbf{0.0007} &  \textbf{0.0281} &  \textbf{0.0456} & \underline{91.01\%} &  \textbf{0.1165} &  \textbf{0.0009} &  \textbf{0.0856} &  \textbf{0.0399} & \underline{92.80\%}\\
			\bottomrule
		\end{tabular}
	}
	\label{tab:overallres}
\end{table*}
\subsection{Setting}
We implement Diff-RNTraj\footnote{\url{https://github.com/wtl52656/Diff-RNTraj}} by PyTorch framework with the Adam optimizer for 30 epochs and a batch size of 256. The learning rate is set to 1e-3 and is halved every 3 epochs. For the denoising network, we set the channel dimensional $C=512$, the positional encoding dimensional $F = 512$, the layer of each block $m=4$ and the kernel size to 3 in dilation convolution with padding followed by~\cite{kong2020diffwave}. We adopt the quadratic schedule for the variance schedule: $\beta_n = (\frac{N-n}{N-1}\sqrt{\beta_1} + \frac{n-1}{N-1}\sqrt{\beta_N})^2$ with the minimum noise level $\beta_1 = 0.0001$ and the maximum noise level $\beta_N = 0.02$. 
\textcolor{black}{We search the number of the diffusion steps $N$ and residual dilation convolution layers $L$ within the parameter space ($N \in \{50,100,200, 500, 1000 \}$, $L \in \{ 4,8,10,15,20 \}$).
To obtain the pre-trained road segment representation \textbf{E} using the Node2vec algorithm, we generate 100 random paths for each node, each with a length of 80. We set the contextual window size of 10, perform a total of 1000 iterations, and search the number of  embedding dimensions $D$ across the parameter space $\{16,32,64,128 \}$.}

Considering that our denoising network can only handle fixed-length shapes in a batch, we initially group trajectories according to their length, and select trajectories of the same length within each batch to train the model. During the sampling stage, we generate an equal number of trajectories at each length as the original dataset.

\subsection{Experiments Results}
We use Diff-RNTraj and several baseline models to generate an equal amount of trajectory data as the original dataset, and we compare the distributions of the generated and original trajectory data across JSD-TD, JSD-SD, JSD-GPD, JSD-RS, and RSC metrics. As depicted in Table~\ref{tab:overallres}, we have the following observations:

Firstly, rule-based methods RWRN and Markov can effectively maintain the spatial validity of trajectories. However, the generated trajectory data struggle to capture real movement patterns, resulting in significant disparities compared to the characteristics of the original trajectories. This indicates that describing complex human travel patterns using simple rules is challenging.

Secondly, the two-stage approaches that first generate GPS points and then perform map matching, show performance improvement compared with rule-based methods. Nonetheless, they still display dissimilarities in characteristics compared to real trajectory data. Furthermore, these methods face difficulties in reliably mapping all GPS points to the road network, achieving only around 82.13\% and 60.85\% in the Porto and Chengdu datasets, making it challenging for further utilization. 

Our Diff-RNTraj achieves the best results across all metrics, with an average improvement of 55.05\% on the Porto dataset and 33.12\% on the Chengdu dataset. Diff-RNTraj notably improves the overall distribution of trajectory data, indicating the effectiveness of the diffusion model in generating trajectories. Additionally, Diff-RNTraj achieves RSC scores surpassing 91\%, demonstrating that our generated trajectory is valid.
\subsection{Ablation Study}
To explore the effectiveness of different components in our Diff-RNTraj, we conduct five ablation experiments on the Porto and Chengdu datasets.
These experiments involve: utilizing binary encoding for discrete roads (Diff-Bit), removing the pre-trained strategy for road segment representation learning (Diff-RNTraj / Pretrain), replacing residual dilation convolution layers with a transformer (Diff-Transformer) and LSTM (Diff-LSTM), and omitting spatial validity loss (Diff-RNTraj / SL).

The experimental results are shown in Table~\ref{tab:overallres}. We observe that when removing the pre-trained strategy, the quality of the generated trajectory decreases, indicating that it is helpful to capture the correlation between road segments.
Diff-Bit has a low RSC, indicating that using binary-coded road segments is inefficient. It easily compromises the spatial validity of trajectories since a single bit error can disrupt the entire trajectory.
The removal of spatial validity loss led to a decrease in RSC metrics on both the Porto dataset (from 91.01\% to 79.68\%) and the Chengdu dataset (from 92.80\% to 87.79\%). This indicates that our spatial validity loss effectively improves the quality of generated trajectories, making them more similar to real trajectories.
The performance is decreased when replacing the residual dilation convolution layer in the denoising network with the Transformer or LSTM. This is because the Diff-Transformer considers the correlation of all trajectory points simultaneously, which introduces additional noise from potentially unrelated distant GPS points, yet Diff-LSTM fails to capture the temporal correlation of future trajectory points. In contrast, the residual dilation convolution layer enables a focus on temporal relationships among nearby trajectory points, which facilitates improved learning of trajectory correlations.
\subsection{Hyperparameter Analysis}
\begin{figure}[t]
	\centering
	\subfigure{
		\includegraphics[width=0.48\linewidth]{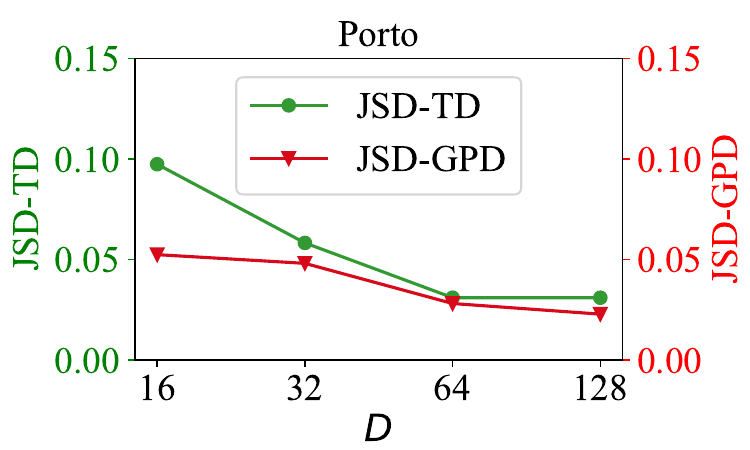}}
	\subfigure{
		\includegraphics[width=0.48\linewidth]{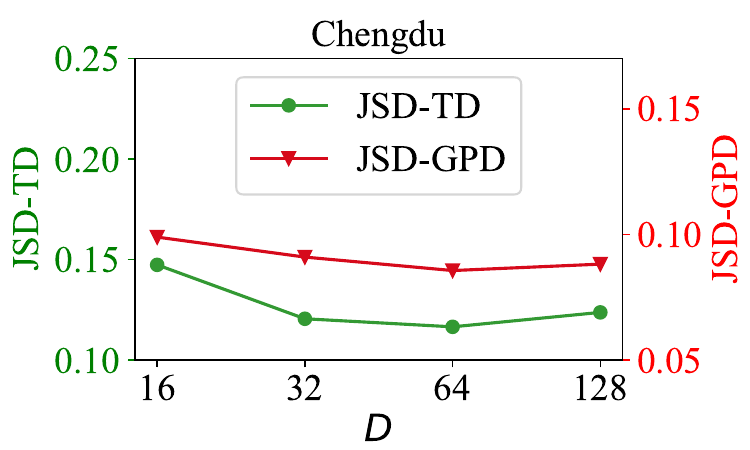}}
	\caption{Hyperparameter analysis of embedding dimensional $D$ of the pre-trained road segment representation.}
	\label{fig:hypermeter_dim}
\end{figure}

\begin{figure}[t]
	\centering
	\subfigure{
		\includegraphics[width=0.48\linewidth]{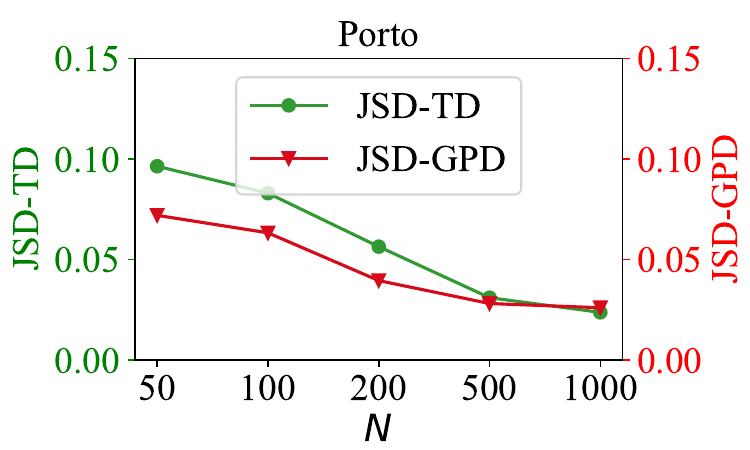}}
	\subfigure{
		\includegraphics[width=0.48\linewidth]{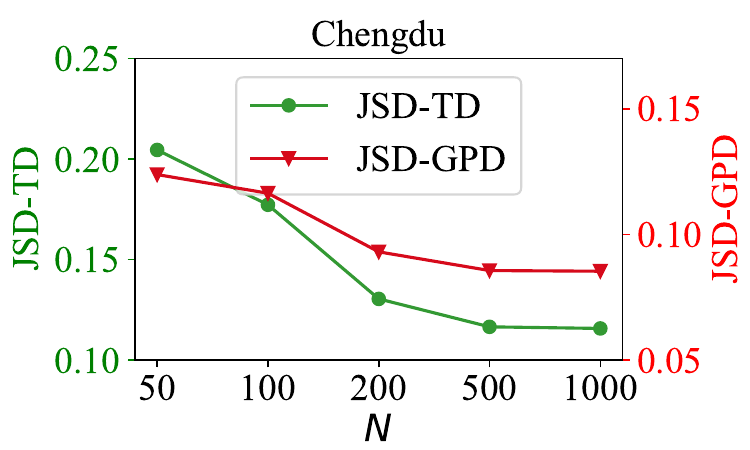}}
	\caption{Hyperparameter analysis of number of diffusion steps $N$.}
	\label{fig:hypermeter_diff_T}
\end{figure}

\begin{figure}[t]
	\centering
	\subfigure{
		\includegraphics[width=0.48\linewidth]{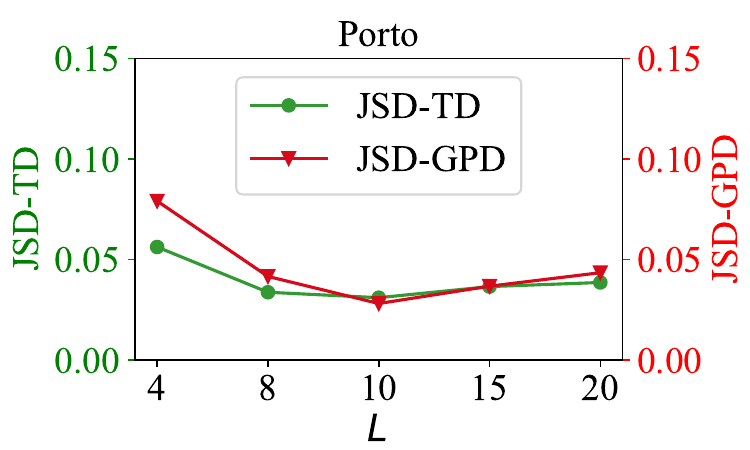}}
	\subfigure{
		\includegraphics[width=0.48\linewidth]{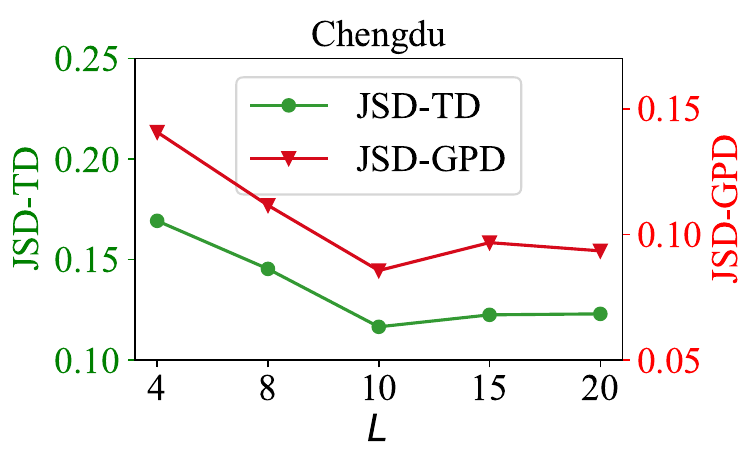}}
	\caption{Hyperparameter analysis of layer of residual dilation convolution layers $L$ in the denoising network.}
	\label{fig:hypermeter_tcn_layer}
\end{figure}

We further explore the impact of various hyperparameters in Diff-RNTraj on the Porto and Chengdu datasets. These hyperparameters include the number of hidden dimensions $D$ in the pre-trained road segment representation, the total diffusion steps $N$, and the number of residual dilation convolution layers $L$.
\begin{itemize}
	\item Pre-trained road segment representation dimension $D$.  We vary $D$ within $\{16, 32, 64, 128\}$ to analyze its impact on trajectory generation. As illustrated in Figure~\ref{fig:hypermeter_dim}, we observe that as $D$ increases, the performance of Diff-RNTraj improves. This suggests that a higher-dimensional space is more effective in capturing road segment information. When $D$ surpasses 64, the performance gains become marginal. Therefore, we choose $D = 64$ for the Porto and Chengdu datasets.
	\item Diffusion steps $N$. As shown in Figure~\ref{fig:hypermeter_diff_T}, we set $N$ to 50, 100, 200, 500, and 1000. We observe that the quality of trajectory generation improves as $N$ increases. When $N$ is greater than 500, the performance gains become less. Therefore, we set $N = 500$ to balance effectiveness and efficiency.
	\item The number of residual dilation convolution layers $L$ in the denoising network. We conduct experiments using various values of $L$, including 4, 8, 10, 15, and 20, to examine how the number of residual dilation convolution layers affects trajectory generation. As depicted in Figure~\ref{fig:hypermeter_tcn_layer}, an increase in $L$ enables trajectory points to better capture correlations with neighboring points, progressively aligning the generated distribution with the actual distribution. However, when $L$ exceeds 10, the generated trajectories deviate from the real distribution. This divergence is caused by the excessive layer stacking in Diff-RNTraj, which leads each trajectory point to incorporate information from distant points, resulting in noise and reduced performance.
\end{itemize}


\subsection{Effectiveness Analysis of the Generated Road Segment Representation}
\begin{figure}[h!]
	\centering
	\subfigure{
		\includegraphics[width=0.48\linewidth]{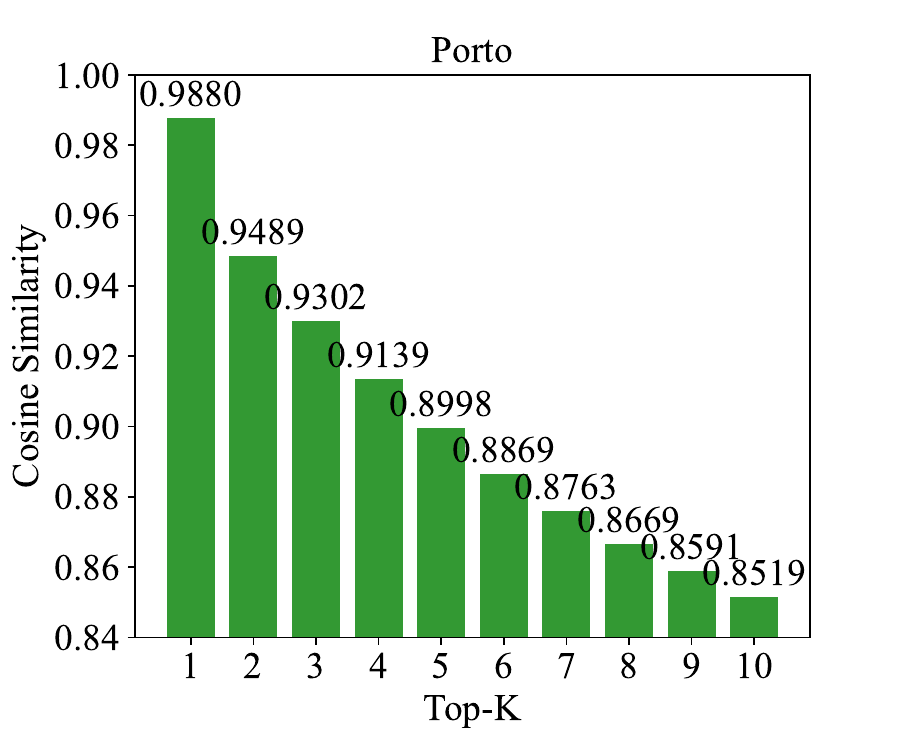}}
	\subfigure{
		\includegraphics[width=0.48\linewidth]{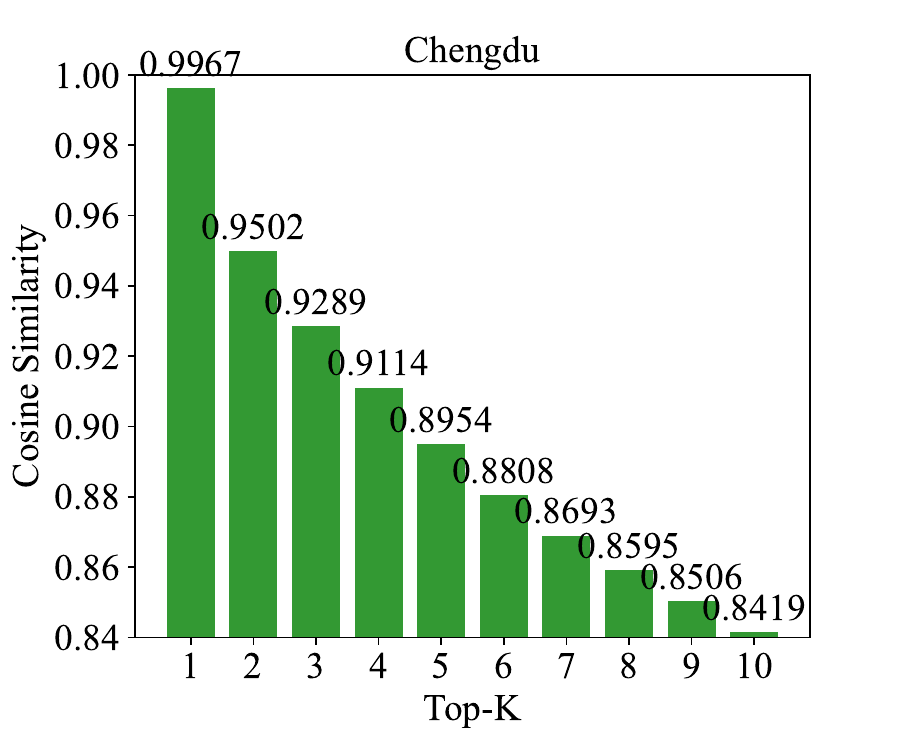}}
	\caption{Average cosine similarity of the generated and pre-trained road segment representation at different top-$k$.}
	\label{fig:robust}
\end{figure}

To assess the effectiveness of our model in generating road segments, we conduct an experiment to check whether our generated road segment representations can be aligned with pre-trained representations.
Specifically, we generate 2,000 trajectories with a length of 20 by $N$ steps reverse denoising process, \textit{i.e.}, we obtain the road segment representations $\hat{X}_e$ for the 40,000 trajectory points.
For each generated road segment representation, we calculate its cosine similarity with the pre-trained road segment representations and record the top 10 most similar road segments. Then, we calculate the average similarity across all trajectory points for different top-$k$ similarity values ($k = 1, \cdots, 10$).

As shown in Figure~\ref{fig:robust}, we observe that it achieves 0.9967 similarity score in the top-1 road segment in the Chengdu dataset. Even in the Porto dataset, which contains 13,696 road segments, the similarity score remains notably high at 0.9880. In contrast, top-2 $\sim$ top-10 road segments exhibit lower similarity, which makes the similar calculation and argmax easy and robust.

\subsection{Visualization Analysis}
To intuitively observe the quality of the generated trajectories, we conduct two visualization experiments, including the trajectory point distributions visualization and the single trajectory visualization.

\begin{figure}[t]
	\centering
	\subfigure[The real trajectories.]{
		\includegraphics[width=0.45\linewidth]{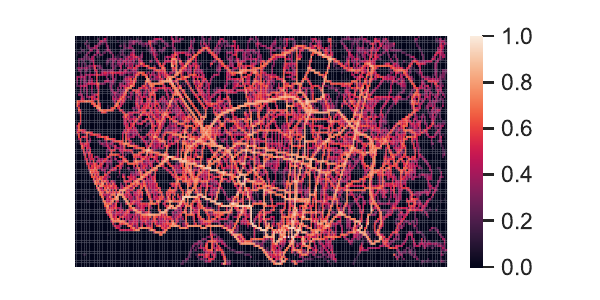}
	}
	\subfigure[The trajectories generated by our model.]{
		\includegraphics[width=0.45\linewidth]{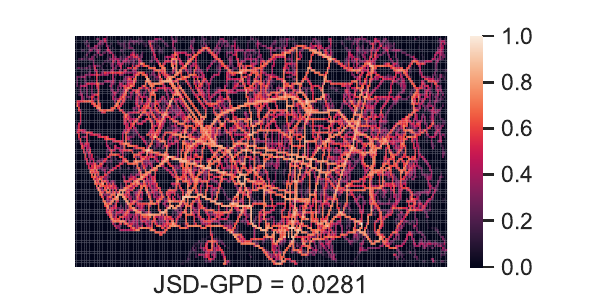}
	}
	\subfigure[The trajectories generated by Diff-Bit.]{
		\includegraphics[width=0.45\linewidth]{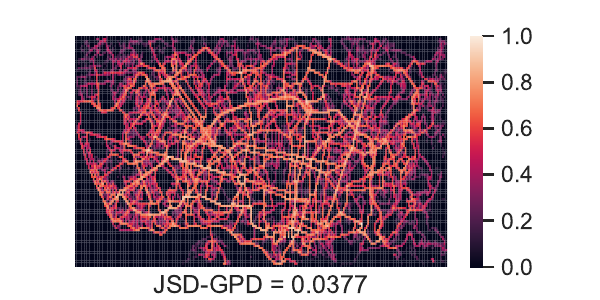}
	}
	\subfigure[The trajectories generated by EETG+HMM.]{
		\includegraphics[width=0.45\linewidth]{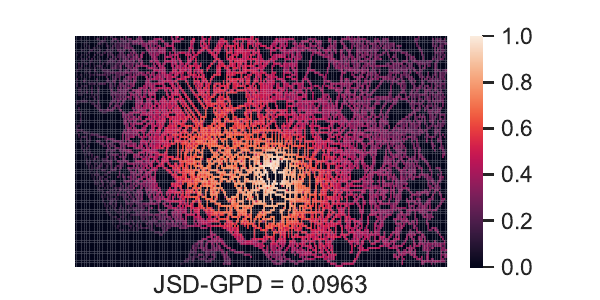}
	}
	\caption{Heatmaps of trajectories in Porto.}
        \label{fig:heatmap_Porto}
\end{figure}
\begin{figure}[t]
	\centering
	\subfigure[The real trajectories.]{
		\includegraphics[width=0.45\linewidth]{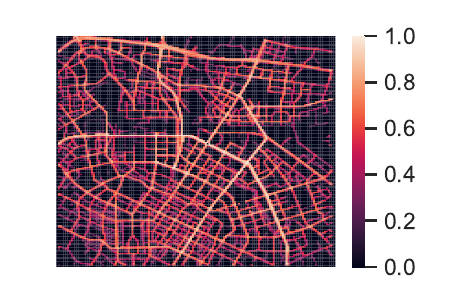}
	}
	\subfigure[The trajectories generated by our model.]{
		\includegraphics[width=0.45\linewidth]{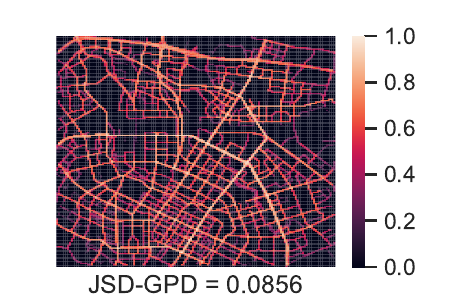}
	}
	\subfigure[The trajectories generated by Diff-Bit.]{
		\includegraphics[width=0.45\linewidth]{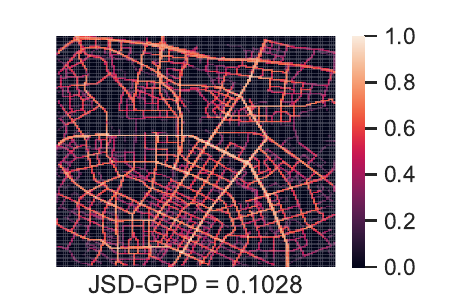}
	}
	\subfigure[The trajectories generated by EETG+HMM.]{
		\includegraphics[width=0.45\linewidth]{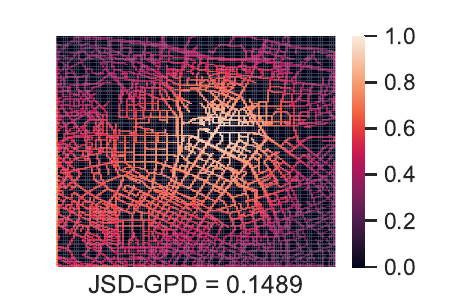}
	}
	\caption{Heatmaps of trajectories in Chengdu.}
        \label{fig:heatmap_CD}
\end{figure}

\subsubsection{Trajectory Point Distribution Visualization} 
In Figures~\ref{fig:heatmap_Porto} and~\ref{fig:heatmap_CD}, we generate an equal number of trajectory points as the original dataset and visualize their availability using the probability density function (PDF). By comparing the heatmaps generated by our Diff-RNTraj, Diff-Bit, EETG+HMM, and real trajectory points, we observe that our model accurately captures the distribution of the trajectory points and reflects the popularity of different roads. 

\begin{figure}
	\centering
	\includegraphics[width=0.5\textwidth]{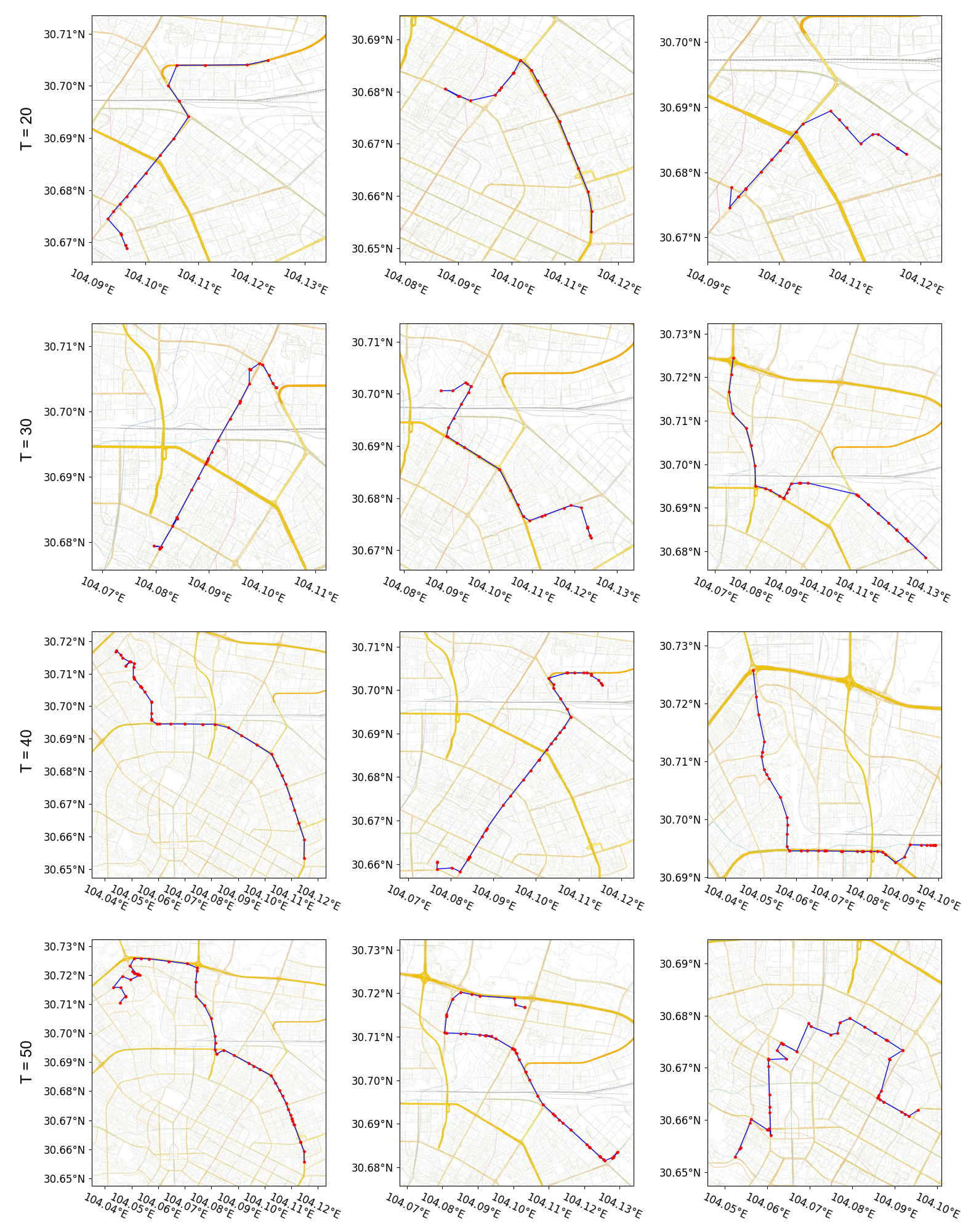}
	\caption{Visualizations of generated trajectories with lengths 20, 30, 40, and 50 in Chengdu.}
	\label{fig:single_traj}
\end{figure} 

\subsubsection{Trajectory Visualization} 
We randomly generate trajectories of different lengths ($T$ = 20, 30, 40, and 50) and plot the GPS points on the road network in Figure~\ref{fig:single_traj}. We observe that these trajectories exhibit patterns consistent with common driving behaviors and show a preference for major roads. Additionally, the vehicles consistently move forward along the roads without any instances of reversing, thus validating that using RNTraj to represent trajectory is suitable.

\subsubsection{\textcolor{black}{Trajectory Diversity Visualization}}
\begin{figure}[t!]
	\centering
\includegraphics[width=0.98\linewidth]{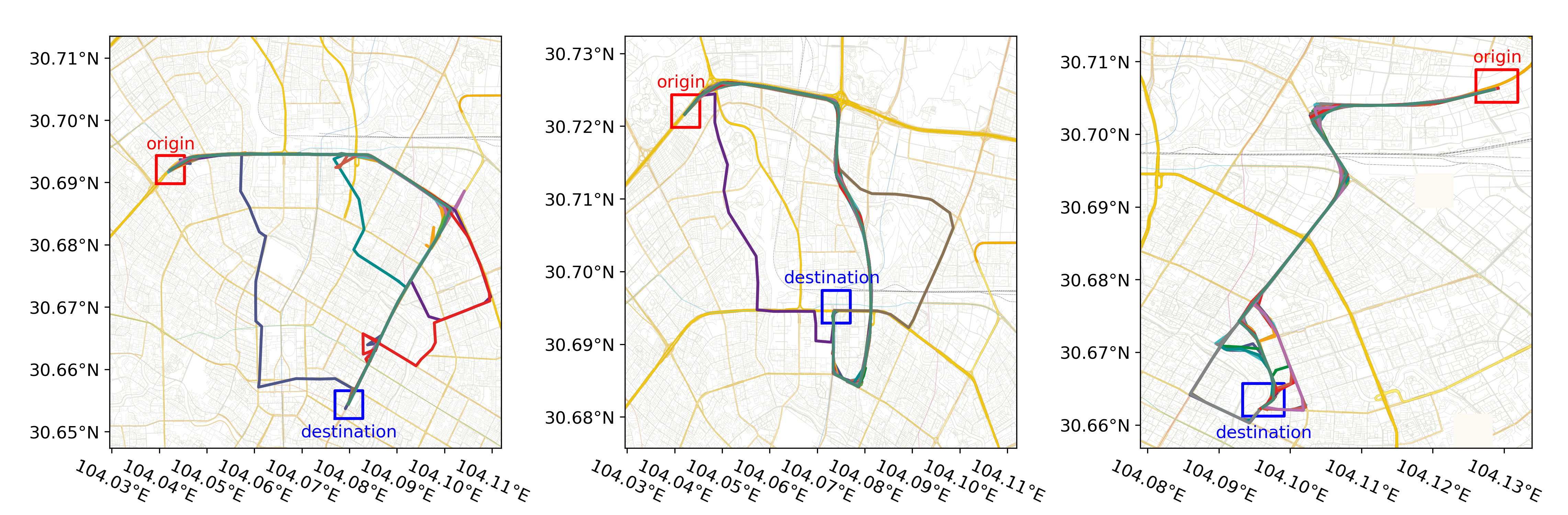}
	\caption{\textcolor{black}{Visualization of trajectory diversity between the same origin and destination regions in Chengdu.}}
	\label{fig:diversity}
 \end{figure}

\textcolor{black}{To explore the diversity of the generated trajectories, we randomly select three origin-destination pairs from the Chengdu dataset and plot the generated. As shown in Figure~\ref{fig:diversity}, our model generated multiple paths for each origin-destination pair, ensuring a level of diversity and avoiding overly deterministic movement patterns. The majority of these trajectories followed the major roads, with transitions to alternative routes as they approached the destination, reflecting common driving behavior.}

\textcolor{black}{However, it is challenging to determine whether certain generated trajectories are abnormal, such as whether the brown trajectory in the middle of Figure~\ref{fig:diversity} represents a detour. This limitation is primarily due to the model's emphasis on modeling the overall distribution and its lack of consideration for conditional information guidance.
}

\begin{table*}[t]	
	\small
	\centering
	\caption{Scalability analysis. The similarity between generated and real trajectory data when using different ratio data to train Diff-RNTraj.}
 \resizebox{1.0\linewidth}{!}{
		\begin{tabular}{c|ccccc|ccccc}
			\toprule
			\multirow{2}{*}{Dataset ratio}&\multicolumn{5}{c|}{Porto} &\multicolumn{5}{c}{Chengdu}\\
			\cmidrule{2-11}
			& JSD-TD & JSD-SD & JSD-GPD & JSD-RS & RSC & JSD-TD & JSD-SD & JSD-GPD & JSD-RS & RSC\\
			\midrule
			25\% & 0.0313 & 0.0009 & 0.0420  & 0.0773 & 89.87\% & 0.1317 & 0.0012 & 0.0930 &  0.0714 & 90.05\% \\
			50\% & 0.0225 & 0.0007& 0.0407 & 0.0536 & 90.26\% & 0.1292 & 0.0011 & 0.0925 &  0.0549 & 91.86\% \\
			75\% & 0.0222 & 0.0007& 0.0366 & 0.0479 &  90.82\% & 0.1174 & 0.0010 & 0.0890 &  0.0456 & 92.20\%\\
			100\% & {0.0221} & {0.0007} & {0.0281} &  {0.0456} & 91.01\% &  {0.1165} &  {0.0009} &  {0.0856} &  {0.0399} & 92.80\%\\
			\bottomrule
		\end{tabular}
		}
	\label{tab:Scability}
\end{table*}

\subsection{Scability Analysis}
In practical situations, collected trajectory data is usually small-scale. To evaluate the model's ability to generate trajectories with limited data, we conduct scalability experiments on two datasets. Diff-RNTraj is trained using 25\%, 50\%, and 75\% of the trajectory data, generating an equivalent amount of trajectory to the original dataset. The generated results are reported in the Table~\ref{tab:Scability}. We observe that even when trained on just 25\% of the data, Diff-RNTraj generates trajectories that closely resemble real data. This demonstrates the effectiveness of our model in scenarios involving small-scale trajectory data, which is more useful in practice.
\begin{figure}[t]
	\centering
\includegraphics[width=0.98\linewidth]{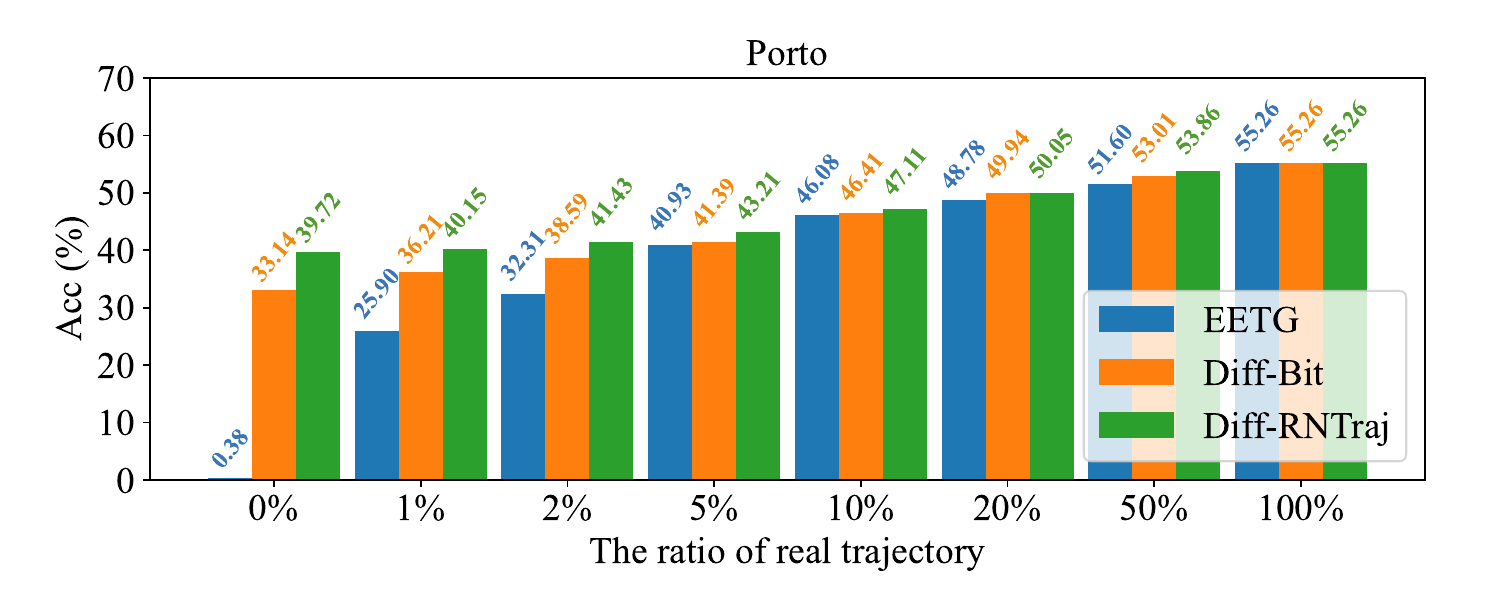}
	\caption{\textcolor{black}{The performance of trajectory prediction on the Porto dataset by mixing real and generated trajectories.}}
	\label{fig:Downstream_Porto}
 \end{figure}
 \begin{figure}[t]
 \centering
\includegraphics[width=0.98\linewidth]{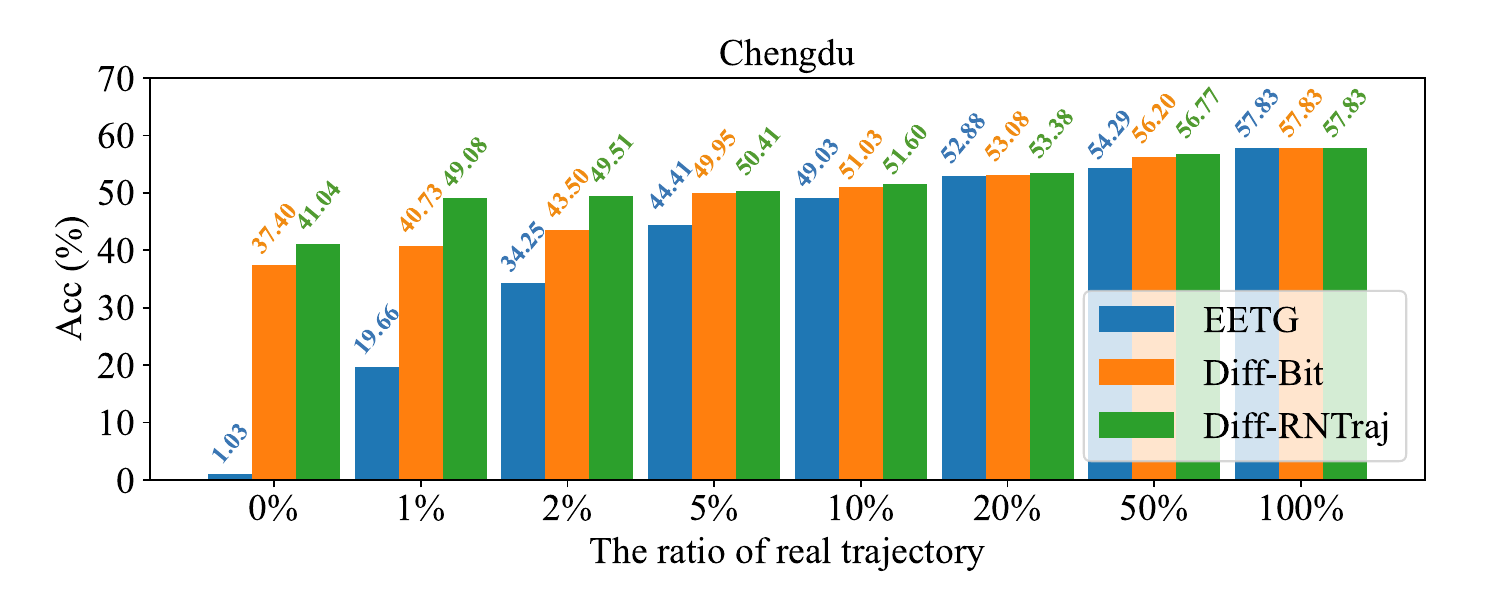}
	\caption{\textcolor{black}{The performance of trajectory prediction on the Chengdu dataset by mixing real and generated trajectories.}}
	\label{fig:Downstream_CD}
\end{figure}
\subsection{Downstream Task}
In this experiment, we take trajectory prediction as the downstream task to evaluate the quality of the generated trajectory. Specifically, we aim to predict the next road segment, given the historical road segment sequence. 
We divide the real trajectory dataset into training, validation, and testing sets with a ratio of 6:2:2. 
In the training set, we keep the overall number of trajectories constant, gradually decrease the ratio of the real trajectories, increase the ratio of the generated trajectories, and train the trajectory prediction model. 
As the ratio of real trajectories decreases, if the prediction performance change is smaller at the test set, it proves that the spatial transfer patterns of the generated trajectories are closer to the real trajectories, and the generated trajectories are more realistic.


Specifically, we choose LSTM as the trajectory prediction model and \textcolor{black}{vary the proportion of real data in the training set to 0\%, 1\%, 2\%, 5\%, 10\%, 20\%, 50\%, and 100\%}. The trajectory prediction results of different generated models are shown in Figure~\ref{fig:Downstream_Porto} and~\ref{fig:Downstream_CD}. 
\textcolor{black}{We observe that Diff-RNTraj demonstrates superior trajectory prediction performance compared to EETG and Diff-Bit, proving that the real trajectories can be replaced by our generated trajectories.
It is worth noting that when the proportion of real data is less than 10\%, or even completely relying on generated results (i.e., with 0\% real data), our model still achieves satisfactory performance, with an average improvement of 3.18\% and 3.81\% on the Porto and Chengdu datasets compared to Diff-Bit. This is because Diff-RNTraj incorporates the pre-trained road segment representation module to model the user's road transfer pattern, enabling the generation of realistic trajectories. This demonstrates the effectiveness of our model for trajectory-based applications.}


\subsection{Computation Cost}
\begin{table}[t]	
	\small
	\centering
	\caption{\textcolor{black}{Computation Cost. NA means the model does not require map-matching.}}
 \resizebox{1.0\linewidth}{!}{
		\begin{tabular}{c|c|c|ccc}
			\toprule
			{Dataset} & \multicolumn{5}{c}{Chengdu / Porto}\\
            \midrule
            \multirow{2}{*}{Methods}& Parameters & Training Time & \multicolumn{3}{c}{Inference Time (seconds / batch)}\\
            \cline{4-6}
            & (million) & (seconds / epoch) & Generation & HMM & Total\\
            \midrule
            TrajVAE + HMM & 12.03 & 217.57 / 269.56 & 2.79 / 2.92 & 128.41 / 182.58 & 131.20 / 185.50 \\
            TrajGAN + HMM & 0.36 & 231.35 / 240.97 & 2.46 / 3.14 & 128.41 / 182.58 & 130.87 / 185.72\\
            EETG + HMM & 8.29 & 327.94 / 404.21 & 6.04 / 6.37 & 128.41 / 182.58 & 134.45 / 188.95\\
            Diff-RNTraj & 27.18 & 194.53 / 226.97 & 13.07 / 13.19 & NA & 13.07 / 13.19\\
			\bottomrule
		\end{tabular}
		}
	\label{tab:cost}
\end{table}
\textcolor{black}{In this section, we conduct computational cost analysis on the Chengdu and Porto datasets using a batch size of 256 with the NVIDIA A40 card. As shown in Table~\ref{tab:cost}, Diff-RNTraj has a larger number of parameters compared to the baselines, as it stacks multiple RDCLs. Despite this, our model's training time is shorter than the baselines. This efficiency is due to RDCL's convolution-based structure, which allows for parallel processing. In contrast, TrajVAE, TrajGAN, and EETG, which are based on the LSTM architecture, process trajectories sequentially.}

\textcolor{black}{For the inference phase, we calculate the time cost of each model to generate a batch of trajectories. When generating trajectories, our model's speed is slower than baselines due to the requirement of 500 iterative denoising steps. However, the baselines necessitate an additional HMM-based map-matching process after trajectory generation, which adds considerable computational expense. This process has a computational complexity of $O(T\mathcal{R}^2)$, where $T$ represents the trajectory length and $\mathcal{R}$ is the number of road segments. Our model, on the other hand, does not require the map-matching step. Consequently, when generating road network-constrained trajectories, the total time of Diff-RNTraj still is the fastest.}

\section{Conclusion}
In this paper, we propose a new problem that directly generates trajectories on the road network. To accomplish this, we utilize a RNTraj of the hybrid format to represent the trajectory and propose a new model called Diff-RNTraj for generating it.
Our approach involves first vectorizing the RNTraj in the continuous space, which serves as the input for the diffusion model. We then develop a RNTraj decoder to map the generated representation from the diffusion model back into RNTraj data. Furthermore, we enhance the validity of the trajectories by introducing a new spatial validity loss at each step of the reverse denoising process.
Extensive experiments conducted on two datasets demonstrate the superiority of our proposed model.

\textcolor{black}{In real scenarios, the drive path of a vehicle usually shows different characteristics under changes in different objective factors, such as traffic conditions and departure time. In the future, we plan to study more specific trajectory generation tasks, \emph{e.g.}, using the different objective factors as conditions to generate trajectories under different patterns.} 


\bibliographystyle{IEEEtran}
\bibliography{main}


\end{document}